\documentclass[journal]{IEEEtran}

\usepackage{cite}
\usepackage{amsmath,amssymb,amsfonts}
\usepackage{algorithmic}
\usepackage{graphicx}
\usepackage{textcomp}

\usepackage{subcaption}

\usepackage{xcolor}
\usepackage{url}
\usepackage[hidelinks]{hyperref}

\usepackage{lscape}

\usepackage{listings}

\definecolor{codegreen}{rgb}{0,0.6,0}
\definecolor{codegray}{rgb}{0.5,0.5,0.5}
\definecolor{codepurple}{rgb}{0.58,0,0.82}
\definecolor{backcolour}{rgb}{0.95,0.95,0.92}

\lstdefinestyle{mystyle}{
    backgroundcolor=\color{backcolour},
    commentstyle=\color{codegreen},
    keywordstyle=\color{magenta},
    numberstyle=\tiny\color{codegray},
    stringstyle=\color{codepurple},
    basicstyle=\ttfamily\footnotesize,
    breakatwhitespace=false,
    breaklines=true,
    captionpos=b,
    keepspaces=true,
    numbers=left,
    numbersep=5pt,
    showspaces=false,
    showstringspaces=false,
    showtabs=false,
    tabsize=2
}

\usepackage{booktabs}
\usepackage{array}
\usepackage{multirow}
\usepackage{rotating}

\usepackage{pifont}
\newcommand{\cmark}{\ding{51}}
\newcommand{\xmark}{\ding{55}}

\usepackage{dsfont}
\usepackage{bm}

\def\BibTeX{{\rm B\kern-.05em{\sc i\kern-.025em b}\kern-.08em
T\kern-.1667em\lower.7ex\hbox{E}\kern-.125emX}}

\begin{document}

\title{BenCoDiC: A Large-Scale Contextual Cyberbullying Corpus
for Bengali Political News Discourse---FULL VERSION}

\author{Md.~Nesarul~Hoque,
        Rudra~Pratap~Deb~Nath,
        Umme~Salma,
        Shima~Chakraborty,
        and~Md.~Hanif~Seddiqui%
\thanks{Md. Nesarul Hoque is with the Big Data, Information and Knowledge
Engineering Lab, Department of Computer Science and Engineering,
University of Chittagong, Chattogram-4331, Bangladesh, and also with the
Department of Computer Science and Engineering, Gopalganj Science and
Technology University, Gopalganj-8105, Bangladesh.}

\thanks{Rudra Pratap Deb Nath, Shima Chakraborty, and Md. Hanif Seddiqui
are with the Big Data, Information and Knowledge Engineering Lab,
Department of Computer Science and Engineering, University of Chittagong,
Chattogram-4331, Bangladesh
(e-mail: rudra@cu.ac.bd).}

\thanks{Umme Salma is with the Department of Computer Science and
Engineering, Bangladesh University, Dhaka-1207, Bangladesh.}

\thanks{This work was partially supported by the Higher Education
Acceleration and Transformation (HEAT) project, funded by the World Bank
and the Government of Bangladesh, through the sub-project no. PIN-13211
and the ICT Division, Bangladesh (Fellowship code number:
1280101-120008431-3821117).}
}

\maketitle


\begin{abstract}
Cyberbullying is a growing concern in online communities, yet existing
Bengali corpora remain limited in contextual information, annotation
transparency, scale, accessibility, and support for multidimensional
analysis, thereby hindering robust detection and comprehensive study of
harasser behaviors and networks. To address these gaps, we introduce the
Bengali Contextual and Diverse Cyberbullying (\textit{BenCoDiC}) corpus,
a publicly available contextual cyberbullying resource for Bengali
political news discourse comprising 50,250 textual instances collected
from 100 YouTube videos and their associated discussions. The corpus
preserves caption--comment--reply relationships and is organized through
a nine-dimensional star schema covering users, videos, textual content,
context, time, engagement, political parties, target identities, and
language categories. Five AI chatbots were initially evaluated for
annotation but did not meet the predefined reliability criterion.
Consequently, a two-tier human annotation process with expert
verification was adopted, achieving Fleiss' $\kappa=0.86$ and
Krippendorff's $\alpha=0.86$. Statistical analyses characterize the
corpus, while transformer-based experiments show that incorporating
conversational context yields modest but consistent improvements over
isolated-text classification. \textit{BenCoDiC} supports context-aware
cyberbullying detection and perpetrator-oriented behavioral and network
analysis. The corpus is available at
\url{https://github.com/bencodic/BenCoDiC-Corpus}.
\end{abstract}



\section{Introduction}
\label{sec:intro}
The rapid spread of affordable Internet and smartphones has made social media (e.g., Facebook, X, YouTube) central to daily communication worldwide. This growth has also intensified cyberbullying, harmful or harassing behavior enabled by online anonymity~\cite{Kim2023Anonymity}. Its consequences are severe, contributing to anxiety, depression, low self-esteem, and suicidal ideation~\cite{Peled2019Cyberbullying}, alongside social isolation and academic decline.

Cyberbullying is a global issue and a critical concern in Bangladesh, where the \textit{Bangladesh Cybercrime Trend 2023} report indicates that cyberbullying-related offenses accounted for 52.21\% of reported online crimes in 2022~\cite{Rahman2023Influence}. Although reporting tools exist, moderating the vast volume of harmful content remains labor-intensive, making automated detection essential~\cite{Jacobs2020Automatic}. Effective systems require benchmark corpora that combine user text with contextual information for deeper analysis of perpetrator behavior and conversational dynamics.

Extensive research in English and other high-resource languages has produced numerous cyberbullying datasets, mostly comment-level, ranging from binary~\cite{Mali2025Automatic, Samee2023Safeguarding} to large multiclass resources~\cite{Founta2018Large, Wang2020SOSNet}, as well as parallel corpora in multiple languages (English and Dutch)~\cite{VanHee2018Automatic}. To overcome limitations of isolated comment-based corpora, session-level datasets incorporating richer context, including videos, images, user metadata, temporal cues, and social-graph features, have also emerged~\cite{Rafiq2015Careful, Hosseinmardi2015Analyzing}.

For Bengali, a low-resource language, available cyberbullying-related datasets remain limited~\cite{Hoque2023Challenges}. Existing resources primarily contain user-generated comments and support either binary classification~\cite{Romim2021Hate,Lora2024BenSarc,Hoque2024Detecting} or multiclass classification~\cite{Ahmed2021Bangla, SihabUsSakib2024Cyberbullying, Hoque2023Leveraging}. Although some studies report structured annotation procedures and reliability measures~\cite{Romim2021Hate,Lora2024BenSarc, Mahmud2024Exhaustive}, several provide limited information on annotation guidelines, annotator qualifications, quality-control procedures, or inter-annotator agreement~\cite{Islam2021Evolutionary, Ahmed2021Bangla, Uddin2024Machine, SihabUsSakib2024Cyberbullying}. Furthermore, some datasets are relatively small~\cite{Jahan2019Abusive, SihabUsSakib2024Cyberbullying, Chakraborty2019Threat} or not publicly accessible~\cite{Ahmed2021Deployment, Jahan2019Abusive, Mahmud2024Exhaustive, SihabUsSakib2024Cyberbullying}, limiting reproducibility, comparative evaluation, and broader downstream use.

Based on these limitations, we identify three key research gaps (RGs) in existing Bengali cyberbullying corpora:

\begin{itemize}
\item[RG1] \textbf{Limited Contextual Information:} Existing corpora commonly contain isolated posts or comments without conversational links or associated metadata, limiting context-aware detection and analyses of user behavior, engagement, and interaction networks.

\item[RG2] \textbf{Insufficiently Documented Annotation:} Many resources inadequately report annotation guidelines, annotator qualifications, quality control, inter-annotator agreement, or evaluations of AI-assisted annotation, raising concerns about label reliability.

\item[RG3] \textbf{Limited Scale and Accessibility:} Many corpora are relatively small or not publicly accessible, restricting reproducibility and downstream use.
\end{itemize}

To address these gaps, we introduce the large-scale \textbf{Ben}gali \textbf{Co}ntextual and \textbf{Di}verse \textbf{C}yberbullying (\textit{BenCoDiC}) corpus for Bengali political news discourse, with the following research contributions (RCs):

\begin{itemize}
\item[RC1] \textbf{Context-Rich Structured Corpus:} BenCoDiC preserves caption--comment--reply relationships and organizes content and metadata through a nine-dimensional star schema, supporting context-aware detection and multidimensional analyses of users, time, engagement, political topics, target identities, and language categories.

\item[RC2] \textbf{Reliable Annotation Framework:} After evaluating five AI chatbots, we adopt a two-tier human annotation process with expert verification, achieving strong agreement (Fleiss’ $\kappa=0.86$, Krippendorff’s $\alpha=0.86$).

\item[RC3] \textbf{Large Open Dataset:} BenCoDiC provides 50,250 labeled samples ($\approx$60\% \textit{Bully}, 40\% \textit{Not Bully}), offering a large publicly available Bengali cyberbullying corpus for political news discourse.
\end{itemize}

The remainder of the paper is structured as follows. \textbf{Section~\ref{sec:related-work}} reviews related work on cyberbullying corpora and highlights key limitations. \textbf{Section~\ref{sec:dataset}} describes the corpus development process and presents detailed corpus analyses, while \textbf{Section~\ref{sec:experiment}} evaluates the impact of incorporating background conversational context. \textbf{Section~\ref{sec:application}} presents potential applications of the corpus. Finally, \textbf{Section~\ref{sec:conclusion}} concludes the paper and outlines future research directions, followed by a discussion of the study limitations.

\section{Related Work}
\label{sec:related-work}

This section reviews cyberbullying corpora across languages. High-resource languages include both large comment-level and context-rich session-level resources, whereas Bengali resources remain more limited in conversational structure, annotation documentation, and accessibility.

\subsection{Cyberbullying Corpora in High-resource Languages}
Significant research in English and other high-resource languages has produced a variety of cyberbullying corpora, mostly at the comment level. For binary classification, Mali et al.~\cite{Mali2025Automatic} introduced a dataset of 30,500 samples, while Founta et al.~\cite{Founta2018Large} released an 80,000-sample corpus annotated across \textit{Normal}, \textit{Spam}, \textit{Abusive}, and \textit{Hateful} categories. Beyond English, Van Hee et al.~\cite{VanHee2018Automatic} constructed English and Dutch corpora (113,698 and 78,387 posts) with detailed guidelines and inter-annotator reliability reporting.

To address the limitations of isolated comments, several works focus on session-level corpora that embed contextual metadata. Rafiq et al.~\cite{Rafiq2015Careful} developed a media-session corpus consisting of videos, comments, and user information to support context-aware cyberbullying detection. Similarly, Hosseinmardi et al.~\cite{Hosseinmardi2015Analyzing} introduced an Instagram session-based dataset combining images, comments, temporal patterns, and social-graph features. More recently, Yang et al.~\cite{Yang2025SCCD} created a Chinese session-based corpus with fine-grained comment-level labels to enable explainable cyberbullying analysis.

In contrast, our study develops a structured Bengali cyberbullying corpus tailored to a low-resource setting. By incorporating background context (caption–comment–reply linkages), user metadata, and content attributes, the corpus supports multiple downstream tasks, including cyberbullying detection, perpetrator identification, and behavioral and network-level analysis.

\subsection{Bengali Cyberbullying and Related Text Resources}

To position BenCoDiC relative to prior Bengali resources, we consider both cyberbullying corpora and related social-media text classification datasets addressing overlapping forms of harmful discourse, including abusive language, hate speech, aggression, and harassment~\cite{Waseem2017Understanding,VanHee2018Automatic}. Although these tasks are related, they are not equivalent and differ in their annotation objectives and label definitions. Accordingly, Table~\ref{tab:cb-corpus-compare} compares their resource characteristics rather than treating their dataset sizes as directly comparable. The table summarizes the resources by \texttt{Reference}, \texttt{Task}, \texttt{Source}, \texttt{Ln} (language code), \texttt{Cls} (classification type), \texttt{CHP} (conversational hierarchy preserved), \texttt{CnM} (contextual metadata), \texttt{CAE} (AI chatbot annotation evaluation), \texttt{AG} (annotation guidelines), \texttt{AP} (annotation process), \texttt{ARM} (annotation reliability measurement), \texttt{Size} (sample count), and \texttt{PA} (public availability). In accordance with the research gaps identified in Section~\ref{sec:intro}, the comparison features are grouped into \texttt{RG1} (\texttt{CHP}, \texttt{CnM}), \texttt{RG2} (\texttt{CA}, \texttt{AG}, \texttt{AP}, \texttt{ARM}), and \texttt{RG3} (\texttt{Size}, \texttt{PA}). A checkmark indicates that the corresponding feature is explicitly reported in the cited study rather than indicating its quality or strength.

\begin{table*}[ht]
	\caption{Comparison of Bengali corpora for cyberbullying and related social-media text classification tasks.}
    \label{tab:cb-corpus-compare}
    \fontsize{8pt}{10pt}\selectfont
	\begin{tabular*}{\textwidth}{
        @{\extracolsep\fill}
        p{2.0cm}p{1.2cm}p{1.2cm}p{1.2cm}p{1.1cm}
        *{2}{>{\centering\arraybackslash}p{0.4cm}}
        *{4}{>{\centering\arraybackslash}p{0.4cm}}
        >{\centering\arraybackslash}p{2.0cm}
        @{}
        >{\centering\arraybackslash}p{0.3cm}
        }
        \toprule%
        \bfseries Reference & \bfseries Task & \bfseries Source & \bfseries Ln & \bfseries Cls & \multicolumn{2}{c}{\bfseries RG1} & \multicolumn{4}{c}{\bfseries RG2} & \multicolumn{2}{c}{\bfseries RG3} \\
        \cmidrule(lr){6-7} \cmidrule(lr){8-11} \cmidrule(lr){12-13}%
         &  &  &  &  & \bfseries CHP & \bfseries CnM & \textbf{CAE} & \bfseries AG & \bfseries AP & \bfseries ARM & \bfseries Size & \bfseries PA \\
         \midrule
        Romim et al.~\cite{Romim2021Hate} & Hate speech & Facebook, YouTube & Bn & BC & \xmark & \xmark & \xmark & \cmark & \cmark & \cmark & 30000 & \cmark \\
        Islam et al.~\cite{Islam2021Evolutionary} & Abusive & Facebook, YouTube & Bn & BC & \xmark & \xmark & \xmark & \xmark & \xmark & \xmark & 12028 & \xmark \\
        Jahan et al.~\cite{Jahan2019Abusive} & Abusive & Facebook & Bn, BnEn, RBn & BC & \xmark & \xmark & \xmark & \xmark & \cmark & \xmark & 2000 & \xmark \\
        Chakraborty and Seddiqui~\cite{Chakraborty2019Threat} & Threat, abusive & Facebook & Bn & BC & \xmark & \xmark & \xmark & \xmark & \xmark & \xmark & 5644 & \xmark \\
        Bhattacharya et al.~\cite{Bhattacharya2020Developing} & Aggression, Misogyny & YouTube & Bn, Hn, En & BC, MC & \xmark & \xmark & \xmark & \cmark & \cmark & \cmark & $>$25,000 & \cmark \\
        Lora et al.~\cite{Lora2024BenSarc} & Sarcasm & Facebook & Bn & BC & \xmark & \xmark & \xmark & \cmark & \cmark & \cmark & 25636 & \cmark \\
        Sihab-Us-Sakib et al.~\cite{SihabUsSakib2024Cyberbullying} & Cyber-bullying & Facebook, YouTube, Instagram & Bn & MC & \xmark & \xmark & \xmark & \xmark & \xmark & \xmark & 2751 & \xmark \\
        Mahmud et al.~\cite{Mahmud2024Exhaustive} & Cyber-bullying & Social media & Bn, Ctg & BC & \xmark & \xmark & \xmark & \cmark & \cmark & \cmark & 5100 (Bn), 5100 (Ctg) & \xmark \\
        Ahmed et al.~\cite{Ahmed2021Bangla} & Cyber-bullying & Facebook & Bn & MC & \xmark & \xmark & \xmark & \xmark & \xmark & \xmark & 44001 & \cmark \\
        Uddin et al.~\cite{Uddin2024Machine} & Cyber-bullying & Facebook & Bn & BC, MC & \xmark & \xmark & \xmark & \xmark & \xmark & \xmark & 44,001~\cite{Ahmed2021Bangla} + 49,999 & \cmark \\
		Ahmed et al.~\cite{Ahmed2021Deployment} & Cyber-bullying & YouTube & Bn, RBn & BC & \xmark & \xmark & \xmark & \xmark & \xmark & \xmark & 5000 (Bn), 7000 (RBn), 12000(Bn, RBn) & \xmark \\
        \midrule
        \textbf{BenCoDiC} (Proposed) & Cyber-bullying & YouTube & \textbf{Bn, RBn, BnEn, En, Emo} & BC & \textbf{\cmark} & \textbf{\cmark} & \textbf{\cmark} & \textbf{\cmark} & \textbf{\cmark} & \textbf{\cmark} & \textbf{50250} & \textbf{\cmark} \\
        \bottomrule
        \multicolumn{13}{p{16.5cm}}{\textbf{Bn} = Bengali; \textbf{RBn} = Romanized Bengali; \textbf{BnEn} = Bengali--English code-mix; \textbf{En} = English; \textbf{Hn} = Hindi; \textbf{Ctg} = Chittagonian; \textbf{Emo} = Emoji-only; \textbf{BC} = Binary Class; \textbf{MC} = Multi-Class.}
	\end{tabular*}
\end{table*}

Most existing resources support binary classification. Islam et al.~\cite{Islam2021Evolutionary} compiled 12,028 abusive samples but reported no annotation details (\texttt{RG2}) and did not publicly release the dataset (\texttt{RG3}). Romim et al.~\cite{Romim2021Hate} published a 30,000-sample hate-speech corpus, while Lora et al.~\cite{Lora2024BenSarc} introduced a 25,636-sample sarcasm dataset; both provide documented annotation procedures and public access. In contrast, Chakraborty and Seddiqui~\cite{Chakraborty2019Threat} developed a smaller corpus of 5,644 Facebook posts and comments with limited annotation documentation. Despite their contributions to Bengali cyberbullying-related research, none of these resources explicitly preserves conversational hierarchy or provides associated contextual metadata (\texttt{RG1}).

Several corpora support multiclass classification. Ahmed et al.~\cite{Ahmed2021Bangla} introduced a 44,001-sample five-class dataset for Bengali online harassment detection, which Uddin et al.~\cite{Uddin2024Machine} later combined with another publicly available 49,999-sample Bengali dataset to construct a 94,000-sample fused resource. Sihab-Us-Sakib et al.~\cite{SihabUsSakib2024Cyberbullying} developed a smaller five-category corpus containing 2,751 samples. These resources do not preserve conversational hierarchy or provide associated contextual metadata (\texttt{RG1}) and provide limited documentation of their annotation procedures (\texttt{RG2}). Moreover, the corpus of~\cite{SihabUsSakib2024Cyberbullying} is relatively small and not publicly accessible (\texttt{RG3}).

Bengali has also been included in multilingual corpora. Mahmud et al.~\cite{Mahmud2024Exhaustive} developed a bilingual Bengali--Chittagonian dataset containing 5,100 samples in each language, whereas Bhattacharya et al.~\cite{Bhattacharya2020Developing} introduced a multilingual aggression and misogyny corpus spanning Bengali, Hindi, and English. Neither resource preserves conversational hierarchy or provides associated contextual metadata (\texttt{RG1}). Bhattacharya et al.~\cite{Bhattacharya2020Developing} provide annotation guidelines, describe the annotation process, and report an annotation reliability measurement. In contrast, the dataset of Mahmud et al.~\cite{Mahmud2024Exhaustive} is not publicly accessible (\texttt{RG3}).

Other corpora incorporate multiple Bengali language forms. Ahmed et al.~\cite{Ahmed2021Deployment} developed Bengali, Romanized Bengali, and combined datasets containing 5,000, 7,000, and 12,000 samples, respectively. However, the resource does not preserve conversational hierarchy or provide associated contextual metadata (\texttt{RG1}), provides insufficient information about annotator qualifications and annotation guidelines (\texttt{RG2}), and is not publicly available (\texttt{RG3}). Similarly, Jahan et al.~\cite{Jahan2019Abusive} compiled a 2,000-sample abusive-language corpus covering Bengali, Romanized Bengali, and Bengali--English code-mixed text. Although its annotation process was described, the guidelines were insufficiently detailed (\texttt{RG2}). The corpus also does not preserve conversational hierarchy or provide associated contextual metadata (\texttt{RG1}) and is not publicly available (\texttt{RG3}).

In summary, existing Bengali cyberbullying-related corpora generally do not preserve conversational hierarchy and provide limited associated contextual metadata (\texttt{RG1}). Many also provide insufficient documentation of annotation practices, while annotation reliability measurements are not consistently reported across studies (\texttt{RG2}). None reports an evaluation of AI chatbot-assisted annotation. In addition, several resources are relatively small or not publicly accessible (\texttt{RG3}). The proposed \textit{BenCoDiC} corpus addresses these limitations by preserving caption--comment--reply relationships, providing structured contextual metadata, documenting and evaluating the annotation process, and releasing a large publicly available dataset (see Table~\ref{tab:cb-corpus-compare}).

\section{BenCoDiC Corpus}
\label{sec:dataset}

\begin{figure}[ht]
\centering
\includegraphics[width=0.45\textwidth]{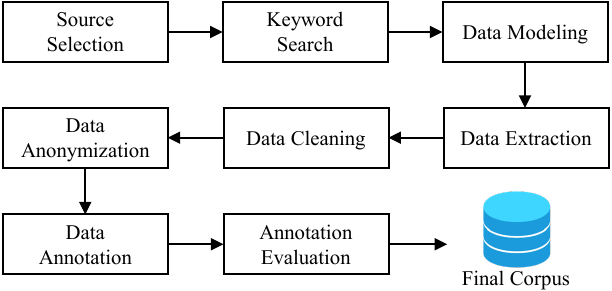}
\caption{Development process of the \textit{BenCoDiC} corpus.}
\label{fig:corpus-development-process}
\end{figure}

This section describes the development of the \textit{BenCoDiC} corpus and presents a statistical analysis of its characteristics. As illustrated in Figure~\ref{fig:corpus-development-process}, the development process comprises source selection, keyword-based video retrieval, data modeling, data extraction, cleaning, anonymization, annotation, and annotation evaluation. The annotated samples are labeled as either \textit{bully} or \textit{not bully}, and the final corpus is constructed after confirming annotation reliability. The remainder of this section describes each development stage and summarizes the key statistical characteristics of the resulting corpus.

\subsection{Source Selection}

YouTube is widely recognized as a major source of offensive and abusive online content~\cite{Islam2021Evolutionary}. During the political unrest in Bangladesh in July--August 2024, bullying content became particularly visible on news-oriented channels, where users interacted through captions, comments, and replies accompanied by contextual and engagement information. We selected the political news domain not merely because of data availability, but because this period generated intensive public discussion involving political leaders, parties, and organizations, providing a socially important setting for examining potentially harmful discourse within preserved conversational relationships. Accordingly, we selected eight widely viewed Bengali news channels: Jamuna TV, SOMOY TV, Channel 24, Ekattor TV, Independent Television, Rtv News, ATN Bangla News, and DBC NEWS. The channels were selected based on subscriber count, number of uploaded videos, total view count, and channel age measured from their YouTube joining dates.

\subsection{Keyword Search}

Because this study focuses on political news in Bangladesh, the selected videos were organized into six categories. Four categories represent major political parties with substantial parliamentary presence since 1991, namely AL, BNP, JaPa, and Jamaat. The National Citizen Party (NCP), formed following the July--August 2024 uprising, constitutes another category. The ``Other'' category comprises parties with parliamentary representation or notable political influence. Relevant videos were identified through a manual keyword search within the eight selected Bengali YouTube news channels using each political party's full name, acronym, and the names of prominent leaders. The search was restricted to videos published between July 24 and November 10, 2024. From the retrieved candidate videos satisfying these criteria, 100 videos were randomly selected while maintaining approximately balanced coverage across both news channels and political categories. The final selection included 12 or 13 videos from each news channel, 16 videos for each of the five political-party categories, and 20 videos in the \textit{Other} category.

\subsection{Data Modeling}
\label{subsec:data-modeling}

\begin{figure}[ht]
	\centering
	\includegraphics[width=0.48\textwidth]{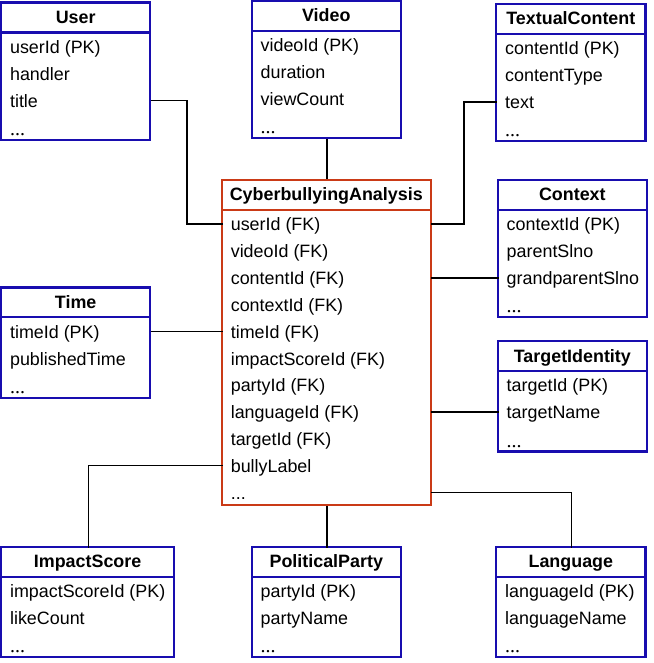}
	\caption{The star schema of the \textit{BenCoDiC} corpus.}
	\label{fig:bencodic-star-schema}
\end{figure}

The \textit{BenCoDiC} corpus is organized using a star schema, as illustrated in Figure~\ref{fig:bencodic-star-schema}, to support multidimensional analysis of cyberbullying content and interactions. At the center of the schema, the \textit{CyberbullyingAnalysis} fact table connects nine dimensions: \textit{User}, \textit{Video}, \textit{TextualContent}, \textit{Context}, \textit{Time}, \textit{ImpactScore}, \textit{PoliticalParty}, \textit{TargetIdentity}, and \textit{Language}. The fact table also contains the \texttt{bullyLabel}, which assigns each textual instance a binary label of either \textit{bully} or \textit{not bully}. The annotation process is described in Section~\ref{subsec:data-annotation}.

The attributes of the fact table and its dimensions are summarized in Table~\ref{tab:column-description}. The \textit{User} dimension stores channel-level metadata, including the channel identifier, handler, title, subscriber count, video count, total view count, and joining date. The \texttt{userId} attribute uniquely represents the YouTube channel associated with the author of a textual instance. For comments and replies, it identifies the corresponding commenter or reply author. During anonymization, a persistent anonymized identifier was assigned so that the same author retains the same anonymized \texttt{userId} across different interactions. This consistency supports user-level aggregation and network analysis without exposing the original identity. The \textit{Video} dimension contains video-level metadata, such as duration, view count, like count, and comment count. The \textit{TextualContent} dimension stores captions, comments, and replies, together with their identifiers, serial numbers, content types, and textual content.

\begin{table*}[ht]
    \centering
	\caption{Attributes and descriptions of the \textit{BenCoDiC} star schema.}
    \label{tab:column-description}%
    \fontsize{8pt}{10pt}\selectfont
	\begin{tabular}{@{}lll@{}}
        \toprule
		\textbf{Fact and} & \textbf{Attribute/} & \textbf{Description} \\
        \textbf{Dimension} & \textbf{Column} &  \\
        \midrule
         & userId & Persistent anonymized identifier of the author's YouTube channel \\
         & handler & Anonymized handler of the author's YouTube channel \\
         & title & Anonymized title of the author's YouTube channel \\
        \textbf{User} & subscriberCount & Total number of channel subscribers \\
         & videoCount & Total number of videos uploaded by the channel \\
         & userViewCount & Total number of views received by the channel's videos \\
         & joinedDate & Date on which the channel joined YouTube \\
         \midrule
         
         & videoId & Unique identifier of a YouTube video \\
         & duration & Duration of the video \\
        \textbf{Video} & viewCount & Total number of views received by the video \\
         & commentCount & Total number of comments posted on the video \\
         & likeCount & Total number of likes received by the video \\
         \midrule

         & contentId & Unique identifier of a caption, comment, or reply \\
        \textbf{TextualContent} & slno & Serial number of a caption, comment, or reply \\
         & contentType & Type of textual content: \textit{caption}, \textit{comment}, or \textit{reply} \\
         & text & Text of a caption, comment, or reply associated with a YouTube video\\
        \midrule
        
         & contextId & Unique identifier of a contextual relationship \\
        \textbf{Context} & parentSlno & Serial number of the parent caption or comment \\
         & grandparentSlno & Serial number of the associated caption serving as the grandparent content \\
         \midrule

        \textbf{Time} & timeId & Unique identifier of a publication date and time \\ 
         & publishedTime & Publication date and time of a video, comment, or reply \\
         \midrule

         & impactScoreId & Unique identifier of an engagement record \\
        \textbf{ImpactScore} & likeCount & Number of likes received by a comment or reply \\
         & replyCount & Number of replies received by a comment \\
         \midrule

         & partyId & Unique identifier of the political party associated with a video \\
        \textbf{PoliticalParty} & partyName & Full name of the political party \\
         & partyShortName & Abbreviated name of the political party \\
         \midrule

         & languageId & Unique identifier of a language category  \\
        \textbf{Language} & languageName & Language category of the textual content: \textit{Bengali}, \textit{Romanized Bengali}, \textit{code-mixed Bengali--English}, \\
         &  & \textit{English}, or \textit{emoji-only} \\
         \midrule

         & targetId & Unique identifier of a target identity category \\
        \textbf{TargetIdentity} &  targetName & Target-identity category: \textit{male}, \textit{female}, \textit{individual}, \textit{people group}, \textit{organization}, or \textit{general people} \\ 
         \midrule
         
        \textbf{Cyberbullying} & bullyLabel & Binary label assigned to the text: \textit{bully} or \textit{not bully} \\
        \textbf{Analysis} &  &  \\
        \bottomrule
	\end{tabular}
\end{table*}

The \textit{Context} dimension preserves the hierarchical relationships among textual instances. Comments are linked to their corresponding captions, whereas replies are linked to both their parent comments and associated captions through the \texttt{parentSlno} and \texttt{grandparentSlno} attributes. Because a caption may receive multiple comments and each comment may receive multiple replies, these relationships enable the reconstruction of threaded interactions.

The \textit{Time} dimension records the publication date and time of videos, comments, and replies. The \textit{ImpactScore} dimension captures engagement information, including like counts for comments and replies and reply counts for comments. The \textit{PoliticalParty} dimension identifies the political party associated with each video. The \textit{Language} dimension records five language categories: \textit{Bengali}, \textit{Romanized Bengali}, \textit{code-mixed Bengali--English}, \textit{English}, and \textit{emoji-only}. Finally, the \textit{TargetIdentity} dimension classifies the target of each textual instance into six categories: \textit{male}, \textit{female}, \textit{individual}, \textit{people group}, \textit{organization}, and \textit{general people}.

By integrating textual content, conversational relationships, channel and video metadata, temporal information, engagement measures, political topics, language categories, and target identities, the schema supports conventional cyberbullying detection as well as ontology- and graph-based analyses, including perpetrator identification, behavioral modeling, network analysis, and link prediction.

\subsection{Data Extraction}
\label{subsec:data-extraction}

Using the keywords described in the preceding subsection, we identified the relevant source videos and retrieved their captions, comments, replies, and associated metadata through the YouTube Data API. Data extraction was implemented using Google Apps Script and divided into two stages to accommodate its execution-time limitation. In the first stage, the textual content and associated video- and interaction-level metadata were extracted for each selected video. In the second stage, channel-level information was retrieved using the channel handler obtained during the first stage. The resulting files were subsequently integrated in Python using the \texttt{handler} field. This process yielded 50,352 textual instances from 100 videos before data cleaning.

Each caption, comment, and reply was stored as a separate record. Their conversational hierarchy was preserved through the \texttt{parentSlno} and \texttt{grandparentSlno} attributes of the \textit{Context} dimension. Channel-level metadata were mapped to the \textit{User} dimension, while video-level attributes were stored in the \textit{Video} dimension. The extracted textual content, publication times, and engagement measures were mapped to the \textit{TextualContent}, \textit{Time}, and \textit{ImpactScore} dimensions, respectively.

Political-party labels were assigned at the video level according to the political party represented directly in the video or through its associated leader or leaders. The assigned party label was then inherited by the corresponding caption, comments, and replies. Target identities were determined using a context-aware, rule-based labeling procedure based on the textual content, available conversational context, and associated metadata. Language categories were assigned using a semi-automated labeling procedure in which initial labels were generated through rule-based identification incorporating the \texttt{langdetect} Python module, followed by manual verification and correction where necessary. These auxiliary dimensions were treated as metadata attributes rather than primary annotation targets and were not subjected to the two-tier inter-annotator agreement procedure used for the binary cyberbullying labels. Consequently, separate reliability coefficients were not computed for these attributes. The resulting records were linked through the \textit{CyberbullyingAnalysis} fact table for subsequent cleaning, annotation, and analysis.

For reproducibility, the accompanying GitHub repository provides the source-video list, corpus schema, extraction procedure, annotation guidelines, data-extraction and cleaning scripts, and a usage example.

\subsection{Data Cleaning}

In the extracted dataset, the \texttt{text} column contains the textual content of captions, comments, and replies. During preprocessing, noisy and irrelevant elements, including HTML tags, hyperlinks, and redundant whitespace, were removed. Samples containing only numeric characters or unsupported non-Bengali or non-Latin scripts were excluded, while emoji-only samples were retained. However, single-word samples were retained because, under context-aware annotation, terms such as \includegraphics[scale=0.30]{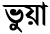} (fake), \includegraphics[scale=0.30]{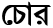} (thief), \includegraphics[scale=0.30]{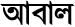} (idiot), \includegraphics[scale=0.30]{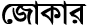} (joker), and \includegraphics[scale=0.30]{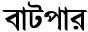} (fraud) can convey explicit bullying. This cleaning process removed 102 noisy samples, reducing the dataset from 50,352 to 50,250 textual instances.

\subsection{Data Anonymization}

To protect user privacy, channel-related identifying information, including user IDs, handlers, and channel titles, was anonymized. The anonymization mapping was kept consistent across the corpus so that the same user retains the same anonymized \texttt{userId}, \texttt{handler}, and \texttt{title} across different comments and replies.

\subsection{Data Annotation}
\label{subsec:data-annotation}
This subsection outlines the data annotation process for the \textit{BenCoDiC} corpus, combining initial experiments with AI chatbots and subsequent human annotation. However, AI chatbots' unreliable performance and inherent limitations led us to adopt human annotation as the primary strategy. Qualified annotators, guided by well-defined instructions and quality checks, ensured reliable labeling, finalized through majority voting to construct the gold-standard corpus.

\subsubsection{Annotation with AI Chatbots}
As a preliminary experiment, we annotated 500 samples using five widely used AI chatbots: ChatGPT, Grok 3, DeepSeek, Gemini, and Perplexity, to evaluate their effectiveness in labeling Bengali texts as either bullying or non-bullying. To ensure fairness, all chatbots in the free-tier version were provided with the same prompt, which required them to assign a binary label while considering the contextual hierarchy of captions, comments, and replies. To examine the stability of their performance, we conducted experiments with four scenarios for each chatbot, varying across different user accounts and days: (Account A, Day X), (Account B, Day X), (Account A, Day Y), and (Account B, Day Y). Each chatbot therefore produced four accuracy scores. The same 500 samples were used in all four scenarios so that variations in accuracy reflected changes in chatbot responses rather than differences in test data. The 90\% threshold was adopted a priori as a conservative study-specific criterion for determining whether chatbot-generated labels were sufficiently reliable for large-scale corpus annotation, rather than as a universal standard for AI annotation reliability. For statistical evaluation, the 90\% threshold was adopted a priori as a conservative study-specific reliability criterion. A chatbot was considered sufficiently reliable only if its annotation accuracy reached at least 90\%.

Table~\ref{tab:ai-chatbot-experiment} reports the accuracy values from the four experiments, along with mean accuracy and 95\% confidence intervals (CIs). The results show that ChatGPT achieved an average accuracy of 72.2\%, Grok 3 reached 79.75\%, DeepSeek 76.0\%, Gemini 70.65\%, and Perplexity 68.55\%. 
Although Grok 3 achieved the highest mean accuracy among the five chatbots, the upper bounds of the corresponding 95\% confidence intervals remained below the predefined 90\% threshold. Therefore, none of the tested free-tier chatbot configurations satisfied the reliability criterion adopted in this study. These results should be interpreted only within the scope of the 500-sample evaluation and the specific free-tier versions tested. Because the four evaluations used the same 500 samples, the reported confidence intervals summarize variation across repeated chatbot runs rather than across independent test samples.

\begin{table}[ht]
    \centering
    \caption{Accuracy and 95\% run-level CIs of free-tier AI chatbots for Bengali cyberbullying annotation.}
    \label{tab:ai-chatbot-experiment}
    \fontsize{8pt}{10pt}\selectfont
    \begin{tabular*}{\columnwidth}{@{\extracolsep\fill}lrrrrrrr}
        \toprule
        \textbf{AI} & \multicolumn{4}{@{}c@{}}{\textbf{Accuracy (\%)}} & \textbf{Mean} & \multicolumn{2}{@{}c@{}}{\textbf{95\% CI (\%)}} \\ 
        \cmidrule{2-5} \cmidrule{7-8}%
        \textbf{Chatbot} & \textbf{E-1} & \textbf{E-2} & \textbf{E-3} & \textbf{E-4} & \textbf{(\%)} & \textbf{LB} & \textbf{UB} \\
        \midrule 
        ChatGPT & 71.00 & 71.40 & 69.80 & 76.60 & 72.20 & 67.41 & 76.99 \\
        Grok 3 & 80.2 & 75.8 & 82.0 & 81.0 & 79.75 & 75.40 & 84.10 \\
        DeepSeek & 76.0 & 75.6 & 78.0 & 74.4 & 76.00 & 73.62 & 78.38 \\
        Gemini & 71.4 & 70.4 & 69.6 & 71.2 & 70.65 & 69.34 & 71.96 \\ 
        Perplexity & 70.6 & 69.6 & 68.0 & 66.0 & 68.55 & 65.35 & 71.75 \\
        \bottomrule
    \end{tabular*}

\end{table}

This experiment was intended as a preliminary assessment rather than a comprehensive evaluation of AI-based annotation methods. Its scope is limited by the use of 500 samples and free-tier chatbot versions. Paid API versions, subsequently updated models, and task-specific fine-tuned language models were not evaluated. Consequently, the findings should not be generalized beyond the tested configurations.

All models responded quickly, although Grok 3’s reasoning mode was occasionally slower. Clear bias patterns were observed: Grok 3 tended to over-predict bully labels, whereas the others leaned toward not bully. Hierarchical labeling remained weak across all systems, particularly for replies. In terms of features, DeepSeek provides a large context window (\textasciitilde128K tokens), Gemini extends up to \textasciitilde1M tokens with regional limitations and free API access, while the remaining chatbots offer smaller or unspecified limits.

Due to the unreliable annotation performance, usage restrictions, labeling bias, and other inherent limitations of the free-tier versions of the five AI chatbots, this study adopts human annotation as the labeling strategy for the entire dataset.

\subsubsection{Human Annotation}
\begin{figure*}[ht]
	\centering
    \includegraphics[width=0.85\textwidth]{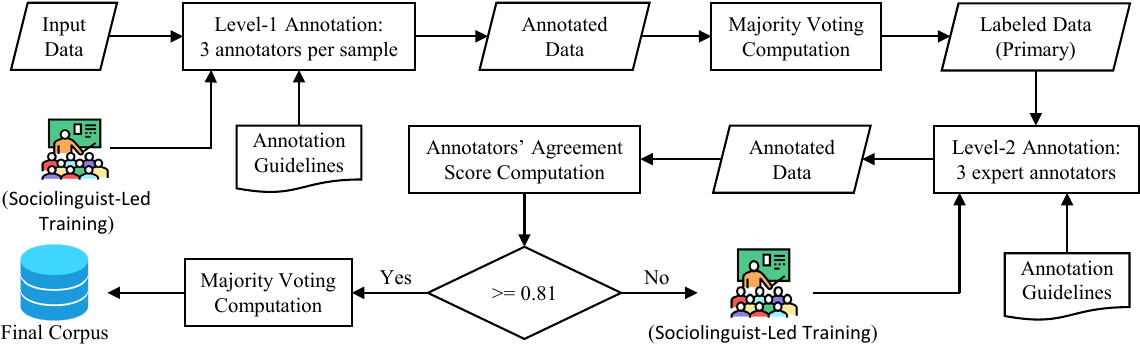}
	\caption{Annotation process of the \textit{BenCoDiC} corpus.}
	\label{fig:annotation-process}
\end{figure*}

A two-tier human annotation strategy was used to construct the gold-standard corpus. As illustrated in Figure~\ref{fig:annotation-process}, the process began with annotator training based on the annotation guidelines. In Level-1, three annotators independently labeled each sample, and majority voting was used to generate the primary labels. These labels were subsequently reviewed in Level-2 by three domain experts. Inter-annotator agreement was then assessed among the Level-2 experts, and the annotations were finalized through majority voting after the predefined reliability threshold of 0.81 was achieved. If the Level-2 agreement score fell below this threshold, the experts received additional guideline-based training before the annotation was reassessed.

\textbf{Annotator Selection and Background:} In Level-1, 51 final-year undergraduate students from the Department of Computer Science and Engineering participated as annotators. Bengali is their first language, while English is their second language. Their education up to the Higher Secondary Certificate, equivalent to Class 12, was conducted primarily in Bengali, while English is the medium of instruction at the undergraduate level. Although they did not have formal academic backgrounds in linguistics or sociology, they were familiar with contemporary Bengali online communication through active use of social-networking platforms such as YouTube and Facebook. No separate pre-annotation screening or qualification test was administered. Instead, their native-language competence, bilingual academic exposure, familiarity with social-media discourse, and subsequent guideline-based training provided the basis for their participation in the Level-1 annotation process. In Level-2, three faculty members from the same department, each with more than five years of research experience in natural language processing and relevant international publications, served as domain experts and independently verified the primary labels.

\textbf{Operational Definition Used in BenCoDiC:}
Cyberbullying overlaps with offensive language, abusive language, toxicity, hate speech, and harassment, but these phenomena are not equivalent because they differ in scope, targets, and annotation objectives~\cite{Waseem2017Understanding,Fortuna2018Survey,VanHee2018Automatic}. Its operationalization also varies across the literature, particularly regarding intent, repetition, and power imbalance, which are difficult to establish reliably in online interactions~\cite{VanHee2018Automatic,Zhang2022Cyberbullying}. Following a less restrictive content-level formulation, BenCoDiC defines a cyberbullying instance as target-directed online content that, within its available conversational context, directly or indirectly threatens, insults, demeans, disparages, or dehumanizes a person, group, or organization, or explicitly endorses such content. Target-directed profanity, slurs, and derogatory emojis are included when they function as such attacks. An identifiable target is therefore required, whereas repetition and power imbalance are not annotation requirements, and annotators do not infer intent beyond the observable textual and contextual evidence. The labels represent cyberbullying content at the contextualized textual-instance level rather than verified longitudinal bullying episodes.

\textbf{Annotation Guidelines and Training:}
A detailed context-aware annotation guideline was developed in consultation with a sociolinguistics expert to support consistent labeling across captions, comments, and replies. The expert conducted three one-hour training sessions to familiarize the annotators with the guideline, contextual interpretation, and the decision criteria for distinguishing between \textit{bully} and \textit{not bully} content. The seven categories were applied as annotation criteria rather than as mutually exclusive class labels. For each textual instance, annotators examined the target text together with its available conversational context and determined whether one or more of the criteria were satisfied. An instance was labeled as \textit{bully} when at least one criterion was satisfied. Otherwise, it was labeled as \textit{not bully}. A single instance could satisfy multiple criteria, and no precedence was imposed among them. Since only the final binary label was retained, inter-category confusion was not assessed as a multiclass annotation outcome. The seven annotation criteria are described below.

\begin{itemize}
    \item[Cat1.] \textit{Profanity or Slurs:} Texts containing target-directed vulgar expressions, profane words, swearing, or slurs are labeled as bullying. For example:
        
    \includegraphics[scale=0.30]{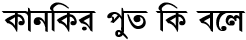} (What does the son of a prostitute say?)

    The word ``\includegraphics[scale=0.30]{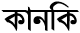}'' (prostitute) functions as a profane and derogatory expression.

    \item[Cat2.] \textit{Direct Personal Attack:} Texts that explicitly threaten, insult, demean, or disparage an individual are labeled as bullying. For example:
    
    \includegraphics[scale=0.30]{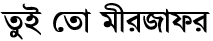} (You are a traitor)

    This comment directly insults the targeted individual.

    \item[Cat3.] \textit{Indirect Personal Attack:} Texts that implicitly mock, provoke, irritate, or disparage an individual without using explicitly aggressive language are also labeled as bullying. For example:
    
    \includegraphics[scale=0.30]{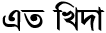} (So much hunger)

    In isolation, the comment may appear harmless. However, when interpreted with the caption \\
    ``\includegraphics[scale=0.30]{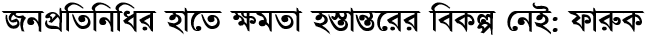}'' (There is no alternative to transferring power to the people's representatives: Faruk), it functions as an indirect personal attack.

    \item[Cat4.] \textit{Direct Group or Organizational Attack:} Texts that explicitly insult, demean, dehumanize, or threaten a group or organization are labeled as bullying. For example:
    
    \includegraphics[scale=0.30]{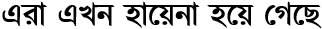} (They have become cruel)

    This comment explicitly dehumanizes the targeted group.

    \item[Cat5.] \textit{Indirect Group or Organizational Attack:} Texts that implicitly mock, provoke, irritate, or disparage a group or organization without explicit abusive language are also labeled as bullying. For example:
    
    \includegraphics[scale=0.30]{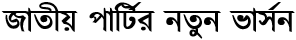} (New version of the National Party)

    When interpreted with the caption \\
    \includegraphics[scale=0.30]{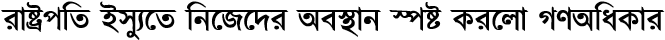} (Gonadhikar clarified its position on the president issue), the comment implicitly disparages the political organization mentioned in the caption.

    \item[Cat6.] \textit{Support for Bullying Statements:} A comment or reply that explicitly endorses a bullying statement is also labeled as bullying. For example:
    
    Comment: \includegraphics[scale=0.30]{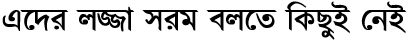} (They have no shame at all)
    
    Reply: \includegraphics[scale=0.30]{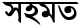} (Agree)

    Although the reply does not independently contain bullying language, it endorses the preceding bullying comment and is therefore labeled as \textit{bully}.

    \item[Cat7.] \textit{Insult through Negative Emojis:} Texts containing derogatory emojis, such as shoe or dog emojis, are labeled as bullying when the surrounding context indicates that they target a person, group, or organization.
\end{itemize}

Several edge cases required careful interpretation. Neutral expressions, agreement replies, emojis, and sarcasm were interpreted using the available conversational context, as illustrated in Cat3, Cat5, Cat6, and Cat7. Romanized Bengali and Bengali--English code-mixed texts posed different linguistic challenges because of non-standard Romanized spelling and language mixing. These variations were considered during interpretation before applying the same seven annotation criteria.

\textbf{Sample Distribution to Annotators:}
The 51 Level-1 annotators were organized into 17 groups of three. Each group was assigned approximately 3,000 of the 50,250 samples, and all three members independently annotated the same subset. For each sample $x_i$, the majority-voted label at either annotation level was computed as

\begin{equation}
\hat{y}_i =
\mathbb{I}\left(
\sum_{j=1}^{3} y_{ij} \geq 2
\right),
\label{eq:majority-voting}
\end{equation}

where $y_{ij}\in\{0,1\}$ denotes the binary label assigned to sample $x_i$ by annotator $j$, with $j\in\{1,2,3\}$ indexing the three annotators, $1$ representing \textit{bully}, and $0$ representing \textit{not bully}. The indicator function $\mathbb{I}(\cdot)$ returns $1$ when at least two of the three annotators assign the \textit{bully} label and returns $0$ otherwise.

At Level-1, Equation~\eqref{eq:majority-voting} was applied to generate the primary labels. The Level-1 majority-voted primary label, together with the corresponding textual and contextual information, was then provided to each of the three Level-2 experts. Thus, the Level-2 experts were not blinded to the Level-1 primary labels. Each expert independently reviewed the samples and assigned a verification label without consulting the other experts. After confirming the required inter-annotator agreement, the same majority-voting rule was applied to the Level-2 labels to determine the final corpus labels.

\subsection{Annotation Assessment}
\label{subsec:annotation-assessment}

Inter-annotator agreement and decision consistency were assessed at both annotation levels. Table~\ref{tab:annotation-assessment} summarizes the agreement scores and the frequencies of unanimous and majority decisions. At Level-1, Fleiss' $\kappa$ and Krippendorff's $\alpha$ were both 0.81 across the complete set of 50,250 samples, while the group-wise agreement scores across the 17 annotator groups ranged from approximately 0.74 to 0.85. Of the Level-1 decisions, 43,286 (86.14\%) were unanimous, where all three annotators assigned the same label, while 6,964 (13.86\%) required a 2--1 majority decision.

\begin{table}[ht]
\centering
\caption{Inter-annotator agreement and decision statistics at Level-1 and Level-2.}
\label{tab:annotation-assessment}
\fontsize{8pt}{10pt}\selectfont
\begin{tabular}{lcc}
\toprule
\textbf{Measure} & \textbf{Level-1} & \textbf{Level-2} \\
\midrule
Fleiss' $\kappa$ & 0.81 & 0.86 \\
Krippendorff's $\alpha$ & 0.81 & 0.86 \\
Unanimous decisions (3--0), $n$ (\%) & 43,286 (86.14) & 45,158 (89.87) \\
Majority decisions (2--1), $n$ (\%) & 6,964 (13.86) & 5,092 (10.13) \\
\bottomrule
\end{tabular}
\end{table}

At Level-2, Fleiss' $\kappa$ and Krippendorff's $\alpha$ both increased to 0.86. The proportion of unanimous decisions increased from 86.14\% to 89.87\%, while the proportion of 2--1 majority decisions decreased from 13.86\% to 10.13\%. Because a 2--1 decision represents disagreement among the three annotators in this binary annotation setting, the raw disagreement rate decreased by 3.73 percentage points from Level-1 to Level-2, corresponding to a relative reduction of 26.88\%.

Comparison of the Level-1 majority-voted primary labels with the final Level-2 majority-voted labels showed that 3,280 samples (6.53\%) changed labels during expert verification. Among all samples, 1,282 (2.55\%) changed from \textit{not bully} to \textit{bully}, while 1,998 (3.98\%) changed from \textit{bully} to \textit{not bully}. These results indicate increased annotation consistency following Level-2 expert verification.

Although disagreement was not separately quantified by content type, language category, or other sample characteristics during annotation, several recurring sources of ambiguity were observed during annotator training and Level-2 verification. These included implicit or context-dependent bullying expressions that appeared non-bullying when read in isolation, benign or politically charged expressions that were sometimes interpreted as bullying, and replies whose interpretation depended strongly on the preceding comment or caption. In such cases, an incorrect interpretation of the surrounding context could also affect the annotation of the associated reply. Romanized Bengali, code-mixed expressions, and emoji-based content also occasionally required additional contextual interpretation. Representative examples of such context-dependent and emoji-based cases are provided in the annotation guidelines above. These observations are qualitative and are therefore not reported as subgroup disagreement rates.

\subsection{Final Corpus}
Following the attainment of high inter-annotator agreement, majority voting was applied to the Level-2 annotations to construct the final \textit{BenCoDiC} corpus, comprising 50,250 samples with twenty-four columns, which are explained in Table~\ref{tab:column-description}. Table~\ref{tab:sample-example} presents sample examples of the cyberbullying classes. The corpus is publicly available at~\url{https://github.com/bencodic/BenCoDiC-Corpus}.

\begin{table}[ht]
	\centering
    \caption{Examples of the cyberbullying classes in BenCoDiC.}
    \label{tab:sample-example}
    \fontsize{8pt}{10pt}\selectfont
	\begin{tabular}{@{}ll@{}}
        \toprule
		\bfseries Text & \bfseries Class \\
        \midrule
        \includegraphics[scale=0.30]{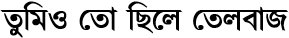} (You were also a sycophant) & Bully \\
		\includegraphics[scale=0.30]{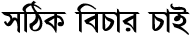} (I want proper justice) & Not Bully \\
        \includegraphics[scale=0.30]{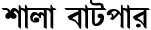} (That bloody cheat!) & Bully \\
        \includegraphics[scale=0.30]{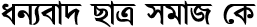} (Thanks to the student community) & Not Bully \\
        \bottomrule
	\end{tabular}
\end{table}

\subsection{Ethical Considerations}
\label{subsec:ethical-considerations}

The development and release of the \textit{BenCoDiC} corpus involved several ethical considerations because the data contain offensive language, political opinions, and potentially sensitive user interactions. Only publicly accessible YouTube content was collected through the YouTube API, and the data were collected and processed exclusively for research purposes. To reduce privacy risks, channel-related identifying information, including user IDs, handlers, and channel titles, was anonymized. No attempt was made to identify or contact users or to infer additional personal information about them.

The corpus preserves caption--comment--reply relationships because conversational context is necessary for context-aware annotation and analysis. These relationships are represented using anonymized internal identifiers, and only metadata relevant to the intended research tasks were retained. The political-party, language, and target-identity attributes describe the contexts of textual instances and should not be interpreted as verified personal characteristics, political affiliations, or demographic profiles of individual users.

Because the corpus contains abusive, profane, and potentially distressing content, annotators received annotation guidelines and training before labeling. To support annotator well-being, they were informed about the sensitive nature of the material and were permitted to pause or withdraw from the annotation process when necessary. Annotation decisions were based on observable textual and conversational evidence rather than assumptions about users' intentions or personal characteristics.

The corpus is intended for research on cyberbullying detection and related behavioral and network analyses. It should not be used for harassment, deanonymization, surveillance, political profiling, or punitive decisions about individuals. Researchers are encouraged to preserve the anonymization measures, avoid attempts to reconstruct user identities, acknowledge potential biases, and interpret findings cautiously. In particular, the corpus represents political discussions collected from selected Bengali news channels during a specific period and should not be considered representative of all Bengali speakers, political communities, or online interactions.

\subsection{Dataset Statistics}
\label{subsec:dataset-stats}

This subsection presents key statistics of the \textit{BenCoDiC} corpus, including class-wise word and sample distributions, content-type proportions, and variations across news channels and political-party categories. It also characterizes the distribution of target identities and language categories. Together, these statistics provide an overview of the corpus and highlight characteristics of cyberbullying within Bengali political news discourse.

\textbf{Text-Length and Lexical Statistics:}
Table~\ref{tab:texual-statistics} summarizes the word- and sample-level characteristics of the corpus. For each class, \textit{minWord}, \textit{maxWord}, and \textit{avgWord} denote the minimum, maximum, and average numbers of words per sample, respectively, while \textit{totalWord} and \textit{totalSample} denote the total numbers of words and samples. On average, \textit{Bully} samples contain more words than \textit{Not Bully} samples (12.52 versus 9.41). Overall, the corpus contains 50,250 samples comprising 566,435 words, with approximately 60\% labeled as \textit{Bully} and 40\% as \textit{Not Bully}.

\begin{table}[ht]
	\centering
    \caption{Textual statistics of the \textit{BenCoDiC} corpus across cyberbullying classes.}
    \label{tab:texual-statistics}
    \fontsize{8pt}{10pt}\selectfont
	\begin{tabular}{@{}lrrrrr@{}}
        \toprule
		\textbf{Class} & \texttt{min} & \texttt{max} & \texttt{avg} & \texttt{total} & \texttt{total} \\
         & \texttt{Word} & \texttt{Word} & \texttt{Word} & \texttt{Word} & \texttt{Sample} \\
        \midrule
		Bully & 1 & 620 & 12.52 & 376,655 & 30,083 \\
		Not Bully & 1 & 629 & 9.41 & 189,780 & 20,167 \\
		Bully+Not Bully & 1 & 629 & 11.27 & 566,435 & 50,250 \\
		\bottomrule
	\end{tabular}
\end{table}

\begin{figure}[ht]
    \centering
    \includegraphics[width=\columnwidth]{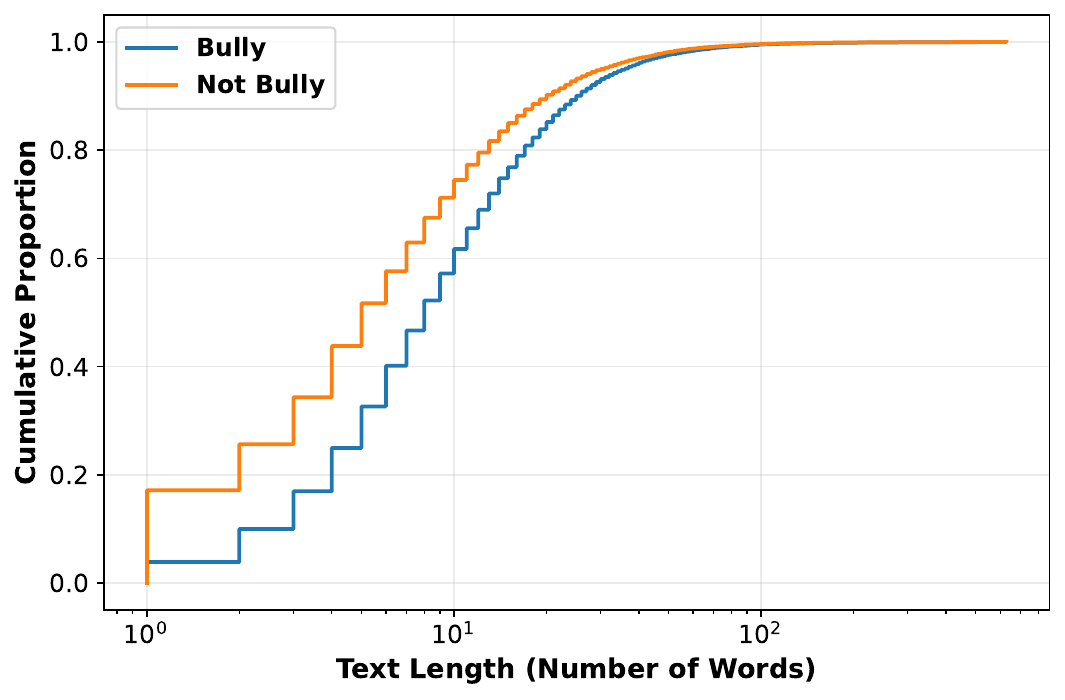}
    \caption{Empirical cumulative distributions of text length across the \textit{Bully} and \textit{Not Bully} classes. The horizontal axis is presented on a logarithmic scale to accommodate the long-tailed word-count distributions.}
    \label{fig:text-length-distribution}
\end{figure}

To further characterize text length, Fig.~\ref{fig:text-length-distribution} presents the empirical cumulative distributions of the number of words per sample. The two classes exhibit largely overlapping distributions, although \textit{Bully} samples tend to be longer. The median text lengths are 8 words for \textit{Bully} and 5 words for \textit{Not Bully} samples.

Table~\ref{tab:lexical-stat} presents additional text-length and lexical characteristics of the two classes. The \textit{Bully} and \textit{Not Bully} classes contain 44,980 and 26,157 unique tokens, respectively. Of the 57,428 unique tokens across both classes, 13,709 are shared, corresponding to a Jaccard vocabulary overlap of 23.87\%. Chi-square analysis further identifies class-associated unigrams, bigrams, and emojis. The \textit{Bully} class shows stronger associations with insulting and politically directed expressions, whereas supportive and affirmative expressions are more strongly associated with the \textit{Not Bully} class. Negative emojis such as \includegraphics[height=1.7ex]{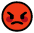}, \includegraphics[height=1.7ex]{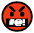}, \includegraphics[height=1.7ex]{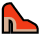}, \includegraphics[height=1.7ex]{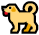}, and \includegraphics[height=1.7ex]{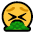} are more strongly associated with \textit{Bully} samples, while \includegraphics[height=1.7ex]{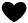}, \includegraphics[height=1.7ex]{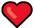}, \includegraphics[height=1.7ex]{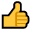}, \includegraphics[height=1.7ex]{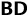}, and \includegraphics[height=1.7ex]{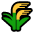} show stronger associations with \textit{Not Bully} samples.

\begin{table*}[ht]
\caption{Additional text-length and class-associated lexical statistics.}
\label{tab:lexical-stat}
\fontsize{8pt}{10pt}\selectfont
\centering
\begin{tabular}{p{3.4cm}p{3.2cm}p{4.1cm}}
\toprule
\textbf{Measure}
& \multicolumn{1}{c}{\textbf{Bully}}
& \multicolumn{1}{c}{\textbf{Not Bully}} \\
\midrule
Median words per sample
& \multicolumn{1}{c}{8}
& \multicolumn{1}{c}{5} \\

Vocabulary size
& \multicolumn{1}{c}{44,980}
& \multicolumn{1}{c}{26,157} \\
\midrule

Shared vocabulary
& \multicolumn{2}{c}{13,709} \\

Combined unique vocabulary
& \multicolumn{2}{c}{57,428} \\

Jaccard vocabulary overlap
& \multicolumn{2}{c}{23.87\%} \\

\midrule
Top five associated unigrams
& \includegraphics[height=1.8ex]{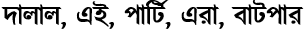}
& \includegraphics[height=1.8ex]{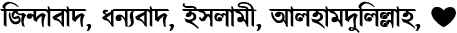} \\

Top five associated bigrams
& \includegraphics[height=1.8ex]{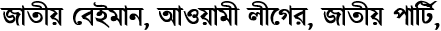} \includegraphics[height=1.8ex]{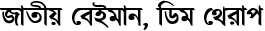}
& \includegraphics[height=1.8ex]{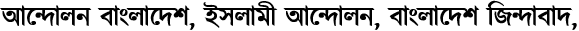} \includegraphics[height=1.8ex]{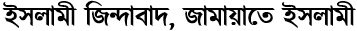} \\

Top five associated emojis
& \includegraphics[height=1.8ex]{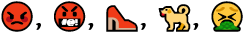}
& \includegraphics[height=1.8ex]{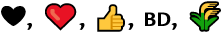} \\

\bottomrule
\end{tabular}
\end{table*}

\begin{figure}[ht]
	\centering
	\includegraphics[width=\columnwidth]{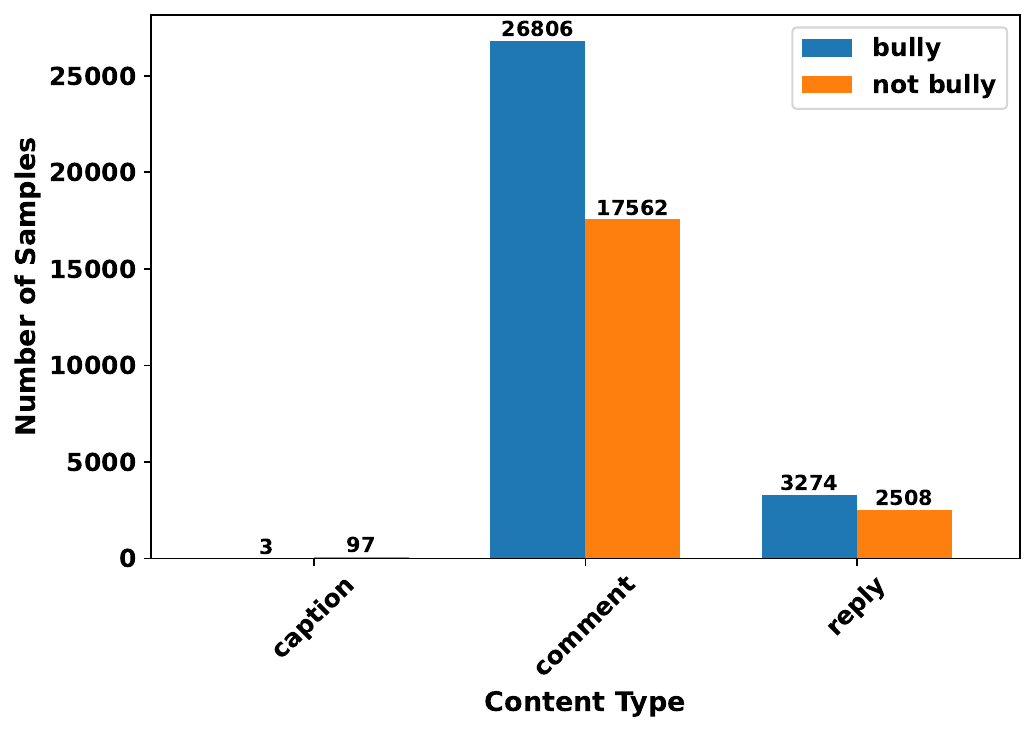}
	\caption{Distribution of \textit{Bully} and \textit{Not Bully} samples across content types.}
	\label{fig:content-type-stat}
\end{figure}

\textbf{Content-Type Statistics:}
Figure~\ref{fig:content-type-stat} presents the distribution of \textit{Bully} and \textit{Not Bully} samples across captions, comments, and replies. Comments constitute the largest portion of the corpus, with 26,806 \textit{Bully} and 17,562 \textit{Not Bully} samples, followed by replies with 3,274 and 2,508 samples, respectively. Captions form only a small portion of the dataset, comprising 3 \textit{Bully} and 97 \textit{Not Bully} samples. Overall, comments account for approximately 88.3\% of the corpus. Since these comments are responses to political-news captions that commonly concern political leaders and parties, the content distribution also reflects the politically focused nature of the collected discourse.
	
\textbf{Engagement Statistics:}
To examine the relationship between cyberbullying and user engagement, we focused on comments, which constitute 44,368 of the 50,250 samples (approximately 88.3\%) and are the only textual content type for which both \textit{likeCount} and \textit{replyCount} are available at the content level. Captions were excluded because their like counts correspond to the associated videos, while replies provide content-level like counts but no corresponding reply-count measure. Comments therefore provide the most consistent basis for jointly analyzing these two forms of engagement.

Table~\ref{tab:engagement-stat} summarizes the engagement characteristics of 26,806 \textit{Bully} and 17,562 \textit{Not Bully} comments. Because the engagement counts are sparse and right-skewed, the mean, median, and interquartile range (IQR) are reported, and Spearman's rank correlation is used as a non-parametric measure of association. \textit{Not Bully} comments received more likes on average than \textit{Bully} comments (3.35 versus 1.62) and slightly more replies (0.18 versus 0.10), although the median for both engagement measures was zero in both classes.

\begin{table*}[ht]
\caption{Engagement characteristics and their association with bullying labels for comments.}
\label{tab:engagement-stat}
\fontsize{8pt}{10pt}\selectfont
\centering
\begin{tabular}{lcccccc}
\toprule
& \multicolumn{2}{c}{\textbf{Bully}} &
\multicolumn{2}{c}{\textbf{Not Bully}} & & \\
\cmidrule(lr){2-3} \cmidrule(lr){4-5}
\textbf{Engagement Measure} &
\textbf{Mean} &
\textbf{Median (IQR)} &
\textbf{Mean} &
\textbf{Median (IQR)} &
\textbf{Spearman's $\rho$} &
$\mathbf{p}$ \\
\midrule
Likes
& 1.62
& 0 (0--0)
& 3.35
& 0 (0--1)
& $-0.1314$
& $<0.001$ \\

Replies
& 0.10
& 0 (0--0)
& 0.18
& 0 (0--0)
& $-0.0628$
& $<0.001$ \\
\bottomrule
\end{tabular}
\end{table*}

With \textit{Not Bully} encoded as 0 and \textit{Bully} as 1, Spearman's correlation was $-0.1314$ for like count and $-0.0628$ for reply count, with both associations statistically significant ($p<0.001$). The negative but small coefficients indicate that higher engagement is weakly associated with \textit{Not Bully} comments in this corpus.

\textbf{Temporal Statistics:}
To examine temporal variation in the corpus, each textual instance was analyzed according to its own publication timestamp rather than the publication date of its associated source video. Although the selected source videos were published from July to November 2024, their associated comments and replies continued to be posted afterward. The observed posting-time range of the complete corpus extends from July 24, 2024 to March 19, 2025, spanning nine calendar months. All 50,250 samples were included in this analysis.

\begin{figure}[ht]
    \centering
    \includegraphics[width=\columnwidth]{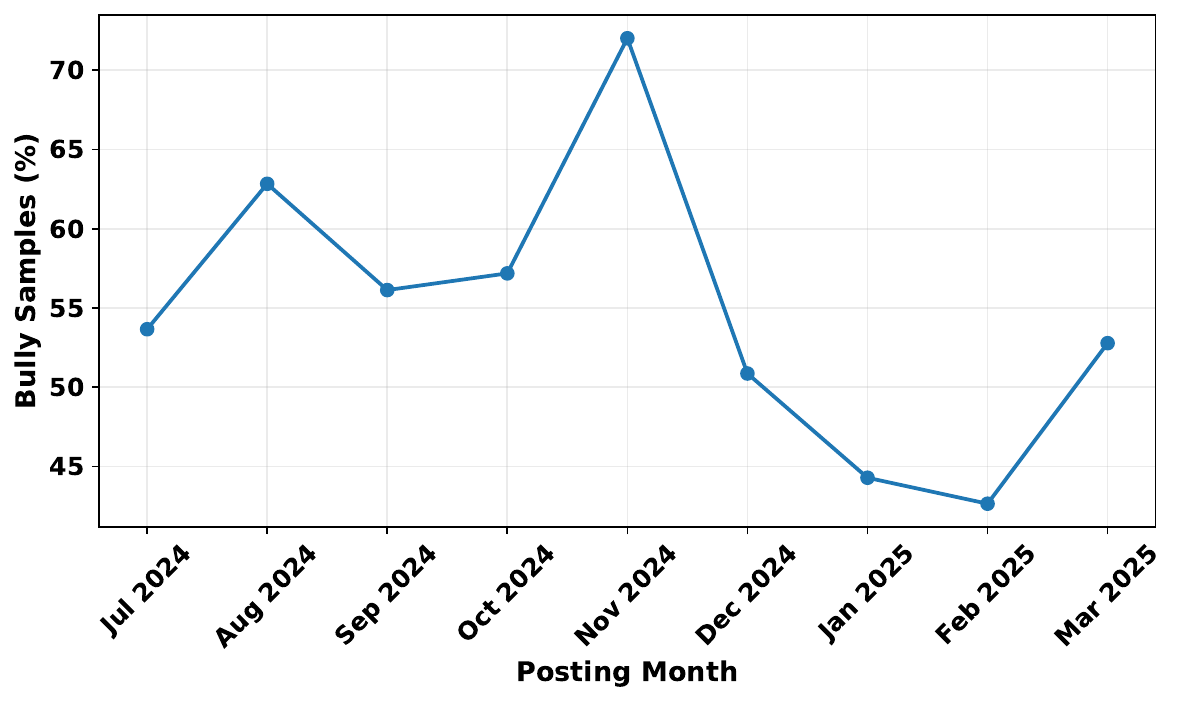}
    \caption{Monthly proportion of \textit{Bully} samples by posting month.}
    \label{fig:temporal-distribution}
\end{figure}

Table~\ref{tab:temporal-stat} summarizes the monthly distribution of labeled samples. Most textual activity occurred from July to November 2024, with August containing the largest number of samples (15,432) and November exhibiting the highest \textit{Bully} proportion (72.01\%). After November, the monthly sample count fell below 120, so the corresponding percentages should be interpreted cautiously. As shown in Figure~\ref{fig:temporal-distribution}, the \textit{Bully} proportion varies over time without a consistent increasing or decreasing trend. July and March represent partial months because the observed corpus begins on July 24, 2024 and ends on March 19, 2025.

\begin{table}[ht]
\caption{Monthly distribution of labeled samples by posting month.}
\label{tab:temporal-stat}
\fontsize{8pt}{10pt}\selectfont
\centering
\setlength{\tabcolsep}{4pt}
\begin{tabular}{lrrrr}
\toprule
\textbf{Month} &
\textbf{Samples} &
\textbf{Bully} &
\textbf{Not Bully} &
\textbf{Bully (\%)} \\
\midrule
Jul. 2024 & 2,831  & 1,519 & 1,312 & 53.66 \\
Aug. 2024 & 15,432 & 9,696 & 5,736 & 62.83 \\
Sep. 2024 & 13,785 & 7,737 & 6,048 & 56.13 \\
Oct. 2024 & 12,849 & 7,347 & 5,502 & 57.18 \\
Nov. 2024 & 5,063  & 3,646 & 1,417 & 72.01 \\
Dec. 2024 & 116    & 59    & 57    & 50.86 \\
Jan. 2025 & 70     & 31    & 39    & 44.29 \\
Feb. 2025 & 68     & 29    & 39    & 42.65 \\
Mar. 2025 & 36     & 19    & 17    & 52.78 \\
\midrule
\textbf{Total} &
\textbf{50,250} &
\textbf{30,083} &
\textbf{20,167} &
\textbf{59.87} \\
\bottomrule
\end{tabular}
\end{table}

\textbf{News-Channel Statistics:}
Table~\ref{tab:news-channel-stat} summarizes the distribution of videos and labeled samples across eight Bengali YouTube news channels. The corpus includes 12 or 13 videos from each channel, indicating relatively even video coverage. However, the number of textual samples varies across channels. Jamuna TV contributes the largest number of samples overall and the highest number of \textit{Bully} instances, followed by Independent Television and Ekattor TV in terms of bullying content. DBC NEWS contributes the fewest samples in both classes. Overall, all eight channels contain substantial numbers of both \textit{Bully} and \textit{Not Bully} samples.

\begin{table}[ht]
	\centering
    \caption{Channel-wise distribution of videos and labeled samples in the \textit{BenCoDiC} corpus.}
    \label{tab:news-channel-stat}
    \fontsize{8pt}{10pt}\selectfont
	\begin{tabular}{@{}lrrrr@{}}
        \toprule
		\textbf{News} & \textbf{Number} & \multicolumn{3}{c}{\textbf{Number of Samples}} \\
        \cmidrule{3-5}
        \textbf{Channel} & \textbf{of Videos} & \textbf{Bully} & \textbf{Not Bully} & \textbf{Total} \\
        \midrule
		Jamuna TV & 12 & 4692 & 2967 & 7659 \\ 
        SOMOY TV & 13 & 3373 & 2780 & 6153 \\ 
        Channel 24 & 13 & 3541 & 2612 & 6153 \\ 
        Ekattor TV & 13 & 4208 & 1954 & 6162 \\ 
        Independent Television & 13 & 4403 & 1956 & 6359 \\ 
        Rtv News & 12 & 3815 & 3091 & 6906 \\
        ATN Bangla News & 12 & 3304 & 3094 & 6398 \\ 
        DBC NEWS & 12 & 2747 & 1713 & 4460 \\
        \midrule
        \textbf{Total} & 100 & 30083 & 20167 & 50250 \\
        \bottomrule
	\end{tabular}
\end{table}

\textbf{Political-Party Statistics:}
Table~\ref{tab:political-party-stat} presents the distribution of videos and labeled samples across the associated political-party categories. AL, BNP, JaPa, Jamaat, and NCP are each represented by 16 videos, while the \textit{Other} category contains 20 videos, providing relatively balanced video coverage across categories. Both \textit{Bully} and \textit{Not Bully} samples occur in every category, although the \textit{Bully} proportion varies considerably, ranging from 43.94\% for Jamaat to 77.70\% for JaPa. This variation further characterizes the politically structured nature of the discourse represented in \textit{BenCoDiC}.

\begin{table}[ht]
	\centering
    \caption{Distribution of videos and labeled samples across political-party categories.}
    \label{tab:political-party-stat}
    \fontsize{8pt}{10pt}\selectfont
	\begin{tabular}{@{}lrrrrr@{}}
        \toprule
		\textbf{Political} & \textbf{Number} & \multicolumn{3}{c}{\textbf{Number of Samples}} & \textbf{Bully} \\ 
        \cmidrule{3-5}
        \textbf{Party} & \textbf{of Videos} & \textbf{Bully} & \textbf{Not Bully} & \textbf{Total} & \textbf{(\%)} \\
        \midrule
		AL     & 16 & 5142 & 2826 & 7968  & 64.53 \\
        BNP    & 16 & 4573 & 3377 & 7950  & 57.52 \\
        JaPa   & 16 & 6404 & 1838 & 8242  & 77.70 \\
        Jamaat & 16 & 3474 & 4432 & 7906  & 43.94 \\
        NCP    & 16 & 4977 & 3087 & 8064  & 61.72 \\
        Other  & 20 & 5513 & 4607 & 10120 & 54.48 \\
        \midrule
        \textbf{Total} & 100 & 30083 & 20167 & 50250 & 59.87 \\
        \bottomrule
	\end{tabular}
\end{table}

\begin{figure}[ht]
	\centering
	\includegraphics[width=\columnwidth]{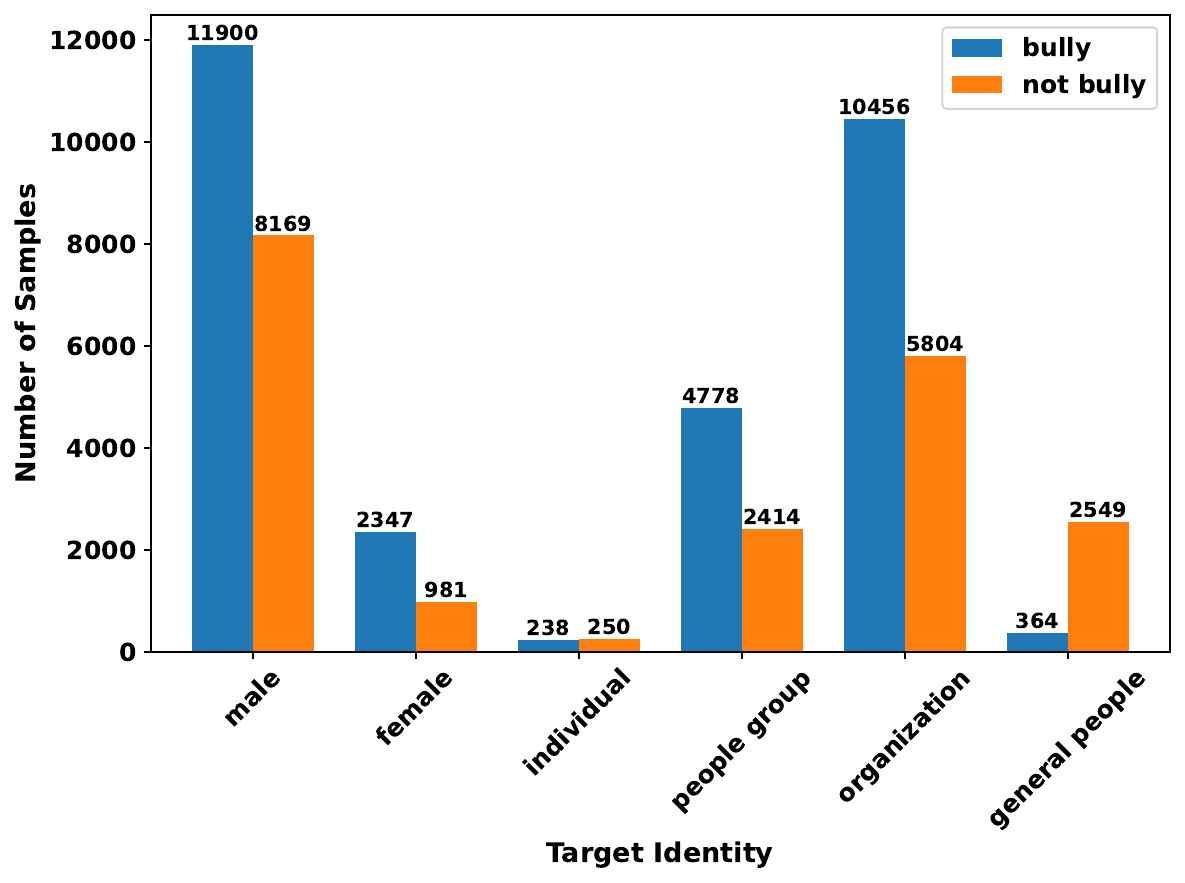}
	\caption{Distribution of \textit{Bully} and \textit{Not Bully} samples across target identities.}
	\label{fig:target-identity-stat}
\end{figure}

\textbf{Target-Identity Statistics:}
Figure~\ref{fig:target-identity-stat} presents the distribution of \textit{Bully} and \textit{Not Bully} samples across six target-identity categories. Among the \textit{Bully} samples, males constitute the largest target category (11,900), followed by organizations (10,456) and people groups (4,778), while female targets account for 2,347 instances. Given that most textual instances are comments on political-news videos, male and female targets frequently correspond to political leaders discussed in the associated captions. Organizations and people groups together account for 50.64\% of all \textit{Bully} samples, indicating that bullying is also frequently directed toward collective political or organizational targets. In contrast, only 12.50\% of instances in the \textit{general people} category are labeled as \textit{Bully}. These patterns further reflect the domain-specific characteristics of political news discourse represented in \textit{BenCoDiC}.

\begin{figure}[ht]
	\centering
	\includegraphics[width=\columnwidth]{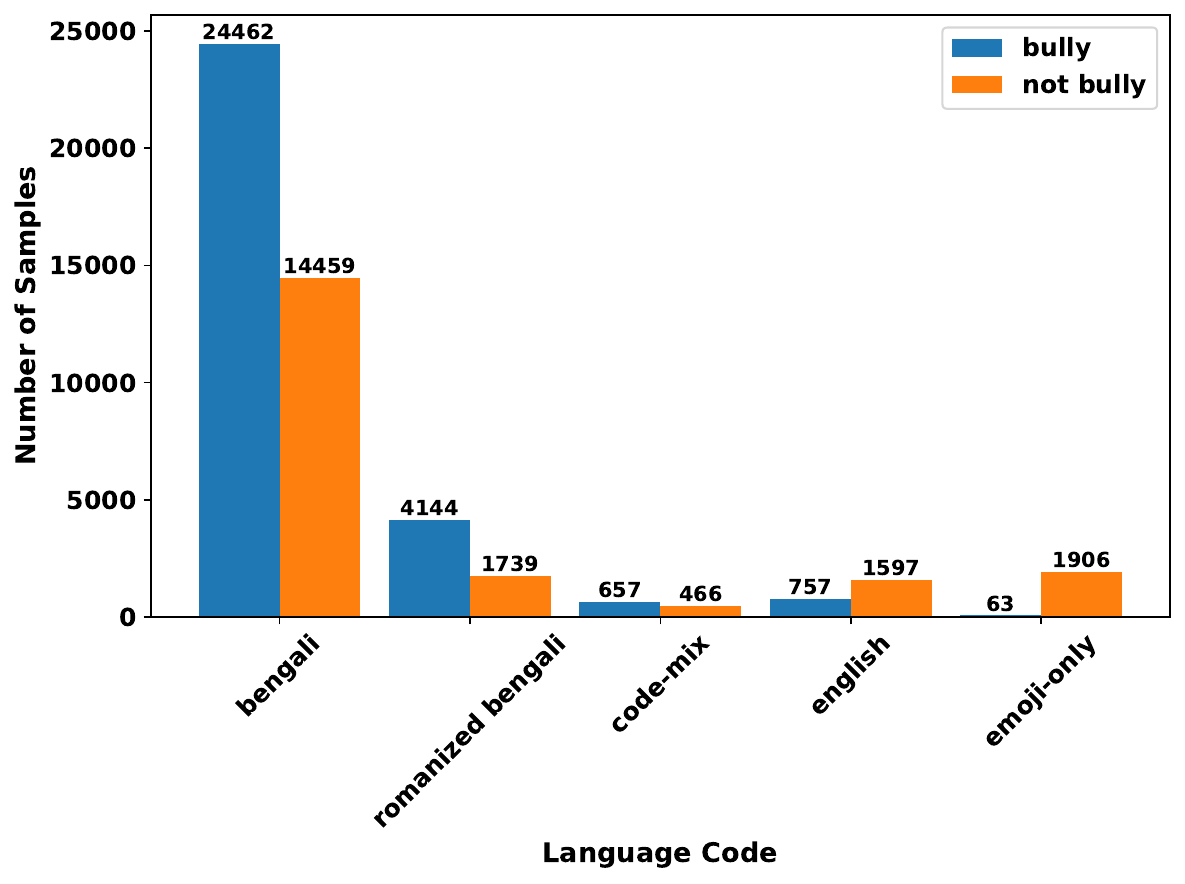}
	\caption{Distribution of \textit{Bully} and \textit{Not Bully} samples across language categories.}
	\label{fig:language-code-stat}
\end{figure}

\textbf{Language Category Statistics:}
Figure~\ref{fig:language-code-stat} presents the distribution of \textit{Bully} and \textit{Not Bully} samples across the five language categories. Bengali accounts for the largest share of the corpus, with 24,462 \textit{Bully} and 14,459 \textit{Not Bully} samples, followed by Romanized Bengali with 4,144 and 1,739 samples, respectively. Code-mixed, English, and emoji-only samples are considerably less frequent. While \textit{Bully} samples outnumber \textit{Not Bully} samples in Bengali, Romanized Bengali, and code-mixed text, the opposite pattern is observed for English and emoji-only content. Overall, the figure reveals a substantial imbalance across the language categories.

\textbf{Target--Language Cross-Tabulation:}
To examine the joint distribution of target identity and language category presented separately in Figures~\ref{fig:target-identity-stat} and~\ref{fig:language-code-stat}, we further analyzed these dimensions across the two cyberbullying classes. Table~\ref{tab:target-language-crosstab} cross-tabulates the six target-identity categories with the five language categories and separately reports the numbers of \textit{Bully} and \textit{Not Bully} samples for each combination.

\begin{table*}[ht]
\caption{Cross-tabulation of target identity and language category across the \textit{Bully} and \textit{Not Bully} classes.}
\label{tab:target-language-crosstab}
\fontsize{8pt}{10pt}\selectfont
\centering
\setlength{\tabcolsep}{2.8pt}
\fontsize{8pt}{10pt}\selectfont
\begin{tabular}{lrrrrrrrrrr}
\toprule
&
\multicolumn{2}{c}{\textbf{Bengali}} &
\multicolumn{2}{c}{\textbf{Romanized Bengali}} &
\multicolumn{2}{c}{\textbf{Code-Mix}} &
\multicolumn{2}{c}{\textbf{English}} &
\multicolumn{2}{c}{\textbf{Emoji-Only}} \\
\cmidrule(lr){2-3}
\cmidrule(lr){4-5}
\cmidrule(lr){6-7}
\cmidrule(lr){8-9}
\cmidrule(lr){10-11}
\textbf{Target Identity} &
\textbf{Bully} & \textbf{Not Bully} &
\textbf{Bully} & \textbf{Not Bully} &
\textbf{Bully} & \textbf{Not Bully} &
\textbf{Bully} & \textbf{Not Bully} &
\textbf{Bully} & \textbf{Not Bully} \\
\midrule
Male
& 9,291 & 5,303
& 2,005 & 721
& 203 & 131
& 356 & 683
& 45 & 1,331 \\

Female
& 1,865 & 655
& 343 & 98
& 64 & 23
& 72 & 88
& 3 & 117 \\

Individual
& 172 & 139
& 56 & 45
& 5 & 4
& 5 & 26
& 0 & 36 \\

People Group
& 4,018 & 2,000
& 579 & 232
& 101 & 56
& 80 & 86
& 0 & 40 \\

Organization
& 8,815 & 4,449
& 1,122 & 403
& 280 & 205
& 229 & 445
& 10 & 302 \\

General People
& 301 & 1,913
& 39 & 240
& 4 & 47
& 15 & 269
& 5 & 80 \\
\bottomrule
\end{tabular}
\end{table*}

The cross-tabulation reveals several patterns that are not evident from the separate target-identity and language-category distributions. Male and organizational targets account for the largest numbers of \textit{Bully} samples, particularly in Bengali and Romanized Bengali content. People-group targets also contain more \textit{Bully} than \textit{Not Bully} samples in Bengali, Romanized Bengali, and Code-Mix content. In contrast, general people contain more \textit{Not Bully} samples across all five language categories. English and Emoji-Only content also contain more \textit{Not Bully} than \textit{Bully} samples across all target-identity categories. These results show that the distribution of cyberbullying labels varies jointly across target identity and language category.

Taken together, the political-party, target-identity, and target--language distributions characterize the domain-specific nature of \textit{BenCoDiC}. The variation in bullying prevalence across political-party categories and the frequent targeting of political figures, organizations, and people groups reflect the political focus of the corpus. The target--language cross-tabulation further shows that these targeting patterns vary across language categories. These characteristics should be considered when comparing \textit{BenCoDiC} with broader social-media datasets. Since all source videos belong to the political-news domain, we do not introduce a separate binary ``high-political-intensity'' flag, which would require an additional operational definition and threshold. Instead, the political-party and target-identity metadata included in the corpus enable users to construct task-specific strata reflecting variations in political focus.

\section{Experimental Evaluation}
\label{sec:experiment}

This section evaluates the effectiveness of BenCoDiC for context-aware cyberbullying classification and examines the contribution of conversational context. The experiments compare context-free and flat-context settings across six transformer-based models using video-disjoint data partitions and repeated evaluations with multiple split and run seeds. Performance is assessed using macro-averaged and weighted metrics, together with paired statistical analysis. We further investigate a simple hierarchical-context formulation, analyze the contribution of individual contextual elements through controlled ablation, and examine representative prediction errors to better understand when contextual information helps or hinders classification.

\subsection{Experimental Protocol}
\label{subsec:exp-protocol}

To obtain a leakage-resistant evaluation of contextual information in BenCoDiC, we adopted a video-disjoint partitioning strategy rather than an instance-level random split. All captions, comments, and replies associated with the same video were assigned exclusively to one partition. This separation is particularly important for the contextual setting, where a comment incorporates its associated caption and a reply incorporates both its caption and parent comment. Consequently, contextual information used in the validation or test set cannot originate from a video represented in the training set.

The 100 videos were divided into training, validation, and test sets using a target ratio of 60:20:20 at the video level. To account for variation arising from data partitioning, three splits were generated using seeds ${0,42,85}$. For each split, 60 videos were assigned to training, 20 to validation, and 20 to testing. Candidate partitions were evaluated according to their agreement with the desired sample proportion, class distribution, and content-type distribution. Each selected split was then verified to ensure no video overlap across partitions and to preserve the caption, comment, and reply hierarchy within each partition.

Table~\ref{tab:video-disjoint-split} summarizes the resulting distributions. Across all three splits, the sample proportions remain close to the intended 60:20:20 ratio, while the proportion of Bully instances ranges from 59.14\% to 61.03\%. The content-type distributions also remain comparable across the corresponding partitions. The exact video assignments for the three benchmark splits are provided in the accompanying GitHub repository.

\begin{table*}[ht]
\centering
\caption{Distribution of the video-disjoint training, validation, and test partitions across three split seeds.}
\label{tab:video-disjoint-split}
\fontsize{8pt}{10pt}\selectfont
\begin{tabular}{clrrrrrrrr}
\hline
\textbf{Split Seed} & \textbf{Partition} & \textbf{Samples} & \textbf{Videos} &
\textbf{Captions} & \textbf{Comments} & \textbf{Replies} & \textbf{Bully} &
\textbf{Not Bully} & \textbf{Bully (\%)} \\
\hline
  & Train      & 30,104 & 60 & 60 & 26,568 & 3,476 & 18,024 & 12,080 & 59.87 \\
0  & Validation & 10,180 & 20 & 20 & 8,888  & 1,272 & 6,133  & 4,047  & 60.25 \\
  & Test       & 9,966  & 20 & 20 & 8,912  & 1,034 & 5,926  & 4,040  & 59.46 \\
\hline
 & Train      & 30,201 & 60 & 60 & 26,617 & 3,524 & 17,879 & 12,322 & 59.20 \\
42 & Validation & 9,952  & 20 & 20 & 8,775  & 1,157 & 6,042  & 3,910  & 60.71 \\
 & Test       & 10,097 & 20 & 20 & 8,976  & 1,101 & 6,162  & 3,935  & 61.03 \\
\hline
 & Train      & 29,934 & 60 & 60 & 26,408 & 3,466 & 17,916 & 12,018 & 59.85 \\
85 & Validation & 10,358 & 20 & 20 & 9,069  & 1,269 & 6,126  & 4,232  & 59.14 \\
 & Test       & 9,958  & 20 & 20 & 8,891  & 1,047 & 6,041  & 3,917  & 60.66 \\
\hline
\end{tabular}
\end{table*}

\subsection{Context Formulation}
\label{subsec:context-formulation}

To evaluate the contribution of conversational context, each textual instance was represented under two settings: context-free and flat context. The prediction target and class label remain identical across both settings, while only the input representation changes. Let $t_i$ denote the target text of instance $i$, $c(i)$ its associated video caption, and $p(i)$ the parent comment when the target is a reply.

In the context-free setting, each caption, comment, or reply is classified independently using only its own text. The input is defined as
\begin{equation}
x_i^{\mathrm{CF}} = t_i.
\end{equation}

In the flat-context setting, contextual information from the caption, comment, and reply hierarchy is combined into a single input sequence. A caption is represented by its own text because no preceding context is available. For a comment, the associated video caption is combined with the target comment. For a reply, the video caption and parent comment are combined with the target reply. Thus,
\begin{equation}
x_i^{\mathrm{FC}} =
\begin{cases}
t_i, & \text{if } t_i \text{ is a caption},\\
c(i) \oplus t_i, & \text{if } t_i \text{ is a comment},\\
c(i) \oplus p(i) \oplus t_i, & \text{if } t_i \text{ is a reply},
\end{cases}
\end{equation}
where $\oplus$ denotes concatenation.

The contextual elements are arranged in conversational order, with the target text placed at the end of the sequence. This paired formulation isolates the effect of background context while keeping the prediction instances, labels, and data partitions unchanged.

\subsection{Models and Training Configuration}
\label{subsec:training-config}

We evaluated six encoder-based transformer models: BanglaBERT, Bangla-Bert-Base, BanglishBERT, IndicBERT v2, mBERT, and XLM-R-base. These models were selected to provide diverse linguistic representations while retaining a common Transformer-based architecture that uses self-attention to capture contextual dependencies in text~\cite{Vaswani2017Attention}. BanglaBERT and Bangla-Bert-Base are Bengali-specific models, BanglishBERT is pretrained on Bengali and English, while IndicBERT v2, mBERT, and XLM-R-base are multilingual models that include Bengali in their language coverage~\cite{Bhattacharjee2022BanglaBERT,Doddapaneni2023Towards,Devlin2019BERT,Conneau2020Unsupervised}. These models have also been employed in Bengali text-classification tasks, including cyberbullying detection, where language-specific and multilingual transformers have demonstrated competitive performance~\cite{SihabUsSakib2024Cyberbullying,Hoque2026Advancing}. Their inclusion enables the effect of conversational context to be examined across different pretraining characteristics and linguistic coverage.

The pretrained checkpoints were
\nolinkurl{csebuetnlp/banglabert} (BanglaBERT),
\nolinkurl{sagorsarker/bangla-bert-base} (Bangla-Bert-Base),
\nolinkurl{csebuetnlp/banglishbert} (BanglishBERT),
\nolinkurl{ai4bharat/IndicBERTv2-MLM-only} (IndicBERT v2),
\nolinkurl{google-bert/bert-base-multilingual-uncased} (mBERT),
and \nolinkurl{xlm-roberta-base} (XLM-R-base).
Each checkpoint was paired with its corresponding pretrained tokenizer.

All six models were fine-tuned under the same experimental configuration to maintain comparability across models and input settings. The learning rate was set to $2\times10^{-5}$ with a weight decay of 0.01, a batch size of 16, and three training epochs. Model selection was based on macro-F1 on the validation set, and the checkpoint with the highest validation macro-F1 was retained for test-set evaluation.

Inputs were tokenized using the corresponding model-specific tokenizer with truncation enabled and a maximum sequence length of 192 tokens. The same tokenization settings were applied to the context-free and flat-context inputs. The maximum length was selected based on the token-length distribution obtained from each model-specific tokenizer. At this threshold, approximately 1.86\% of contextual inputs exceeded the limit for mBERT, while the corresponding proportion remained below 0.31\% for each of the other five models.

To account for variability arising from both data partitioning and model fine-tuning, experiments were conducted using three video-disjoint split seeds $\{0,42,85\}$. For each split, every model-setting combination was independently trained using three run seeds $\{7,12,51\}$. The context-free and flat-context settings used the same split and run seed combinations, yielding nine matched evaluations for each model.

\subsection{Evaluation Metrics and Statistical Protocol}
\label{subsec:metrics-stat-protocol}

Performance was evaluated using macro precision, macro recall, macro F1-score, weighted F1-score, and accuracy. Given the moderately imbalanced class distribution of BenCoDiC, macro-averaged metrics were emphasized because they assign equal importance to both classes. Macro F1-score was treated as the primary evaluation metric, while weighted F1-score and accuracy were reported as complementary measures of overall classification performance.

For each model and input setting, performance was summarized as the mean and standard deviation across the nine evaluations obtained from the combinations of data splits and training runs. This reporting strategy captures performance variation associated with both data partitioning and model fine-tuning.

To assess the effect of conversational context, the context-free and flat-context results were compared using matched split and run combinations. For each split $j$ and run $r$, the difference in macro F1-score was calculated as

\begin{equation}
\Delta F1_{j,r} =
F1^{\mathrm{FC}}_{j,r} - F1^{\mathrm{CF}}_{j,r},
\end{equation}

where $\mathrm{FC}$ and $\mathrm{CF}$ denote the flat-context and context-free settings, respectively. Positive values indicate improved performance after incorporating conversational context. The mean of the paired differences was used to quantify the overall contextual gain.

To evaluate the uncertainty associated with this gain, a 95\% paired confidence interval (CI) was estimated from the matched differences. The interval was calculated as

\begin{equation}
\overline{\Delta F1}
\pm
t_{0.975,8}
\frac{s_{\Delta}}{\sqrt{9}},
\end{equation}

where $\overline{\Delta F1}$ denotes the mean paired difference, $s_{\Delta}$ denotes sample standard deviation of the paired differences, and $t_{0.975,8}$ is the critical value of the $t$ distribution with eight degrees of freedom. If the entire 95\% confidence interval lies above zero, it provides evidence that the flat-context setting achieves a positive mean improvement over the context-free setting.

\subsection{Comparison of Context-Free and Flat-Context Results}
\label{subsec:flat-context-comparison}

Table~\ref{tab:context-performance} presents the performance of the six transformer models under the context-free and flat-context settings. The flat-context setting improves the mean macro precision, macro recall, macro F1-score, weighted F1-score, and accuracy for every model. Among the evaluated models, BanglaBERT achieves the highest overall performance. Its macro F1-score increases from $0.8277 \pm 0.0115$ in the context-free setting to $0.8353 \pm 0.0085$ with flat context, while its accuracy increases from $0.8335 \pm 0.0121$ to $0.8426 \pm 0.0080$. IndicBERT v2 shows closely comparable performance, achieving a macro F1-score of $0.8347 \pm 0.0107$ and an accuracy of $0.8419 \pm 0.0099$ with flat context. The consistent direction of improvement across all six models indicates that preceding conversational information provides useful evidence beyond the target text alone.

\begin{table*}[ht]
\centering
\caption{Performance comparison of the context-free and flat-context settings. Results are reported as mean $\pm$ standard deviation across nine evaluations.}
\label{tab:context-performance}
\fontsize{8pt}{10pt}\selectfont
\setlength{\tabcolsep}{4pt}
\begin{tabular}{llccccc}
\hline
\textbf{Model} &
\textbf{Setting} &
\textbf{Macro Precision} &
\textbf{Macro Recall} &
\textbf{Macro F1} &
\textbf{Weighted F1} &
\textbf{Accuracy} \\
\hline

BanglaBERT
& Context-free
& $0.8259 \pm 0.0128$
& $0.8315 \pm 0.0086$
& $0.8277 \pm 0.0115$
& $0.8343 \pm 0.0115$
& $0.8335 \pm 0.0121$ \\

& Flat context
& $\mathbf{0.8359 \pm 0.0092}$
& $\mathbf{0.8351 \pm 0.0087}$
& $\mathbf{0.8353 \pm 0.0085}$
& $\mathbf{0.8425 \pm 0.0078}$
& $\mathbf{0.8426 \pm 0.0080}$ \\
\hline

Bangla-Bert-Base
& Context-free
& $0.7691 \pm 0.0158$
& $0.7681 \pm 0.0141$
& $0.7684 \pm 0.0148$
& $0.7786 \pm 0.0138$
& $0.7788 \pm 0.0142$ \\

& Flat context
& $0.7758 \pm 0.0135$
& $0.7717 \pm 0.0172$
& $0.7732 \pm 0.0156$
& $0.7838 \pm 0.0134$
& $0.7847 \pm 0.0125$ \\
\hline

BanglishBERT
& Context-free
& $0.8163 \pm 0.0138$
& $0.8219 \pm 0.0132$
& $0.8184 \pm 0.0138$
& $0.8253 \pm 0.0132$
& $0.8245 \pm 0.0134$ \\

& Flat context
& $0.8249 \pm 0.0107$
& $0.8258 \pm 0.0118$
& $0.8251 \pm 0.0109$
& $0.8325 \pm 0.0102$
& $0.8324 \pm 0.0101$ \\
\hline

IndicBERT v2
& Context-free
& $0.8253 \pm 0.0129$
& $0.8314 \pm 0.0114$
& $0.8276 \pm 0.0126$
& $0.8341 \pm 0.0124$
& $0.8334 \pm 0.0127$ \\

& Flat context
& $0.8350 \pm 0.0096$
& $0.8345 \pm 0.0120$
& $0.8347 \pm 0.0107$
& $0.8419 \pm 0.0101$
& $0.8419 \pm 0.0099$ \\
\hline

mBERT
& Context-free
& $0.7915 \pm 0.0127$
& $0.7984 \pm 0.0114$
& $0.7938 \pm 0.0126$
& $0.8013 \pm 0.0122$
& $0.8001 \pm 0.0125$ \\

& Flat context
& $0.8009 \pm 0.0106$
& $0.8038 \pm 0.0105$
& $0.8022 \pm 0.0105$
& $0.8102 \pm 0.0098$
& $0.8098 \pm 0.0099$ \\
\hline

XLM-R-base
& Context-free
& $0.8171 \pm 0.0116$
& $0.8202 \pm 0.0096$
& $0.8183 \pm 0.0105$
& $0.8256 \pm 0.0103$
& $0.8252 \pm 0.0106$ \\

& Flat context
& $0.8255 \pm 0.0097$
& $0.8236 \pm 0.0122$
& $0.8244 \pm 0.0109$
& $0.8322 \pm 0.0100$
& $0.8325 \pm 0.0097$ \\
\hline
\end{tabular}
\end{table*}

The magnitude and uncertainty of the contextual improvement are further examined in Table~\ref{tab:context-significance}. The mean gain in macro F1-score ranges from 0.0048 to 0.0084, corresponding to approximately 0.48 to 0.84 percentage points. The largest mean improvement is obtained by mBERT, followed by BanglaBERT and IndicBERT v2. Although these gains are modest in magnitude, the 95\% paired confidence intervals lie entirely above zero for all six models. Under the paired statistical protocol described in Section~\ref{subsec:metrics-stat-protocol}, these results are considered statistically significant and provide evidence of a positive mean effect of incorporating conversational context across the repeated evaluations.

\begin{table}[ht]
\centering
\caption{Paired analysis of the contextual gain in macro F1-score.}
\label{tab:context-significance}
\fontsize{8pt}{10pt}\selectfont
\setlength{\tabcolsep}{4pt}
\begin{tabular}{lccc}
\hline
\textbf{Model} &
\textbf{Mean $\Delta$F1} &
\textbf{95\% CI} &
\textbf{Significant} \\
\hline
BanglaBERT
& 0.0076
& $[0.0042,\ 0.0109]$
& Yes \\

Bangla-Bert-Base
& 0.0048
& $[0.0027,\ 0.0069]$
& Yes \\

BanglishBERT
& 0.0067
& $[0.0040,\ 0.0094]$
& Yes \\

IndicBERT v2
& 0.0070
& $[0.0038,\ 0.0103]$
& Yes \\

mBERT
& 0.0084
& $[0.0053, 0.0114]$
& Yes \\

XLM-R-base
& 0.0061
& $[0.0038,\ 0.0085]$
& Yes \\
\hline
\end{tabular}
\end{table}

Overall, the results demonstrate that incorporating caption and parent-comment information yields small but systematic improvements across models with different linguistic and pretraining characteristics. These findings support the usefulness of the conversational structure provided by BenCoDiC and motivate a more detailed examination of the contribution of individual context elements in the following subsection.

\subsection{Hierarchical Context and Context-Element Ablation Analysis}
\label{subsec:hierarchical-ablation}

To examine whether preserving the conversational components separately provides an effective alternative to flat concatenation, we implemented a simple hierarchical-context baseline. BanglaBERT was selected as the representative encoder because it achieved the highest overall macro F1-score among the six models in the preceding context comparison. In the hierarchical formulation, the available contextual components and the target text $t_i$, which may be a caption, comment, or reply, are represented separately. The video caption, parent comment, and target text are encoded using a shared BanglaBERT encoder $E$. Let

\begin{equation}
h_c = E(c(i)), \qquad
h_p = E(p(i)), \qquad
h_t = E(t_i),
\end{equation}

where $h_c$, $h_p$, and $h_t$ denote the representations of the caption, parent comment, and target text, respectively. For a caption target, only $h_t$ is available. For a comment target, $h_c$ and $h_t$ are available, while for a reply target, all three representations are available. The representations are concatenated as

\begin{equation}
h_i = h_c \oplus h_p \oplus h_t,
\end{equation}

where $\oplus$ denotes representation concatenation. When a contextual component is unavailable, its corresponding representation is replaced by a zero vector of the same dimension. The resulting $3d$-dimensional representation is passed through dropout and a linear classification layer to predict the class label.

The hierarchical-context setting achieved a macro F1-score of $0.8320 \pm 0.0114$, a weighted F1-score of $0.8385 \pm 0.0108$, and an accuracy of $0.8379 \pm 0.0112$. These results lie between the context-free and flat-context performance of BanglaBERT reported in Table~\ref{tab:context-performance}. The context-free setting achieved a macro F1-score of $0.8277 \pm 0.0115$, whereas the flat-context setting achieved $0.8353 \pm 0.0085$. The same overall pattern is observed for weighted F1-score and accuracy. Thus, the hierarchical formulation improves over target-only classification but does not surpass flat-context integration. This suggests that conversational context is beneficial, while direct sequence-level integration of contextual information is more effective than the simple representation-level fusion considered in the hierarchical baseline.

To further identify the contribution of individual contextual elements, we conducted a controlled ablation analysis within the same hierarchical architecture. The analysis focuses on replies because they are the only content type for which both contextual sources, the video caption and parent comment, are available. Evaluating the ablation variants only on replies therefore enables the effects of the two contextual elements to be compared without the results being dominated by captions or comments.

Four configurations were considered:

\begin{align*}
A0 &: \mathbf{0} \oplus \mathbf{0} \oplus h_t, \\
A1 &: h_c \oplus \mathbf{0} \oplus h_t, \\
A2 &: \mathbf{0} \oplus h_p \oplus h_t, \\
A3 &: h_c \oplus h_p \oplus h_t,
\end{align*}

where $\mathbf{0}$ denotes a zero vector corresponding to an unavailable contextual component. A0 represents the target-only control within the hierarchical architecture, A1 incorporates the caption, A2 incorporates the parent comment, and A3 incorporates both contextual elements. A3 corresponds to the full hierarchical-context setting described above. All four configurations retain the same shared encoder, $3d$-dimensional fusion representation, classifier, training configuration, data splits, and run seeds. Thus, only the availability of the contextual elements varies across the ablation settings.

Table~\ref{tab:context-element-ablation} presents the reply-level results averaged across the nine evaluations. A0 obtains a macro F1-score of $0.7789 \pm 0.0260$. Incorporating the caption in A1 increases the mean macro F1-score to $0.7817 \pm 0.0204$, while incorporating the parent comment in A2 yields a higher score of $0.7855 \pm 0.0223$. The full hierarchical configuration, A3, achieves the highest reply-level macro F1-score of $0.7882 \pm 0.0186$.

\begin{table*}[ht]
\centering
\caption{Reply-level context-element ablation results for BanglaBERT. Results are reported as mean $\pm$ standard deviation across nine evaluations.}
\label{tab:context-element-ablation}
\footnotesize
\setlength{\tabcolsep}{5pt}
\begin{tabular}{lccccc}
\hline
\textbf{Configuration} &
\textbf{Caption} &
\textbf{Parent Comment} &
\textbf{Macro F1} &
\textbf{Weighted F1} &
\textbf{Accuracy} \\
\hline
A0 & No  & No  & $0.7789 \pm 0.0260$ & $0.7821 \pm 0.0280$ & $0.7813 \pm 0.0275$ \\
A1 & Yes & No  & $0.7817 \pm 0.0204$ & $0.7848 \pm 0.0216$ & $0.7837 \pm 0.0214$ \\
A2 & No  & Yes & $0.7855 \pm 0.0223$ & $0.7888 \pm 0.0234$ & $0.7878 \pm 0.0233$ \\
A3 & Yes & Yes & $\mathbf{0.7882 \pm 0.0186}$ & $\mathbf{0.7917 \pm 0.0198}$ & $\mathbf{0.7906 \pm 0.0197}$ \\
\hline
\end{tabular}
\end{table*}

Relative to A0, adding caption context in A1 improves the mean reply macro F1-score by approximately 0.28 percentage points, whereas adding the parent comment in A2 yields a larger improvement of approximately 0.66 percentage points. Combining both contextual elements in A3 produces the largest mean gain of approximately 0.93 percentage points over A0. These results indicate that both contextual sources contribute useful information for reply classification, with the parent comment showing a greater individual contribution than the video caption. The highest performance obtained by A3 further suggests that combining immediate conversational context with the broader caption context provides the most informative representation among the evaluated hierarchical variants.

\subsection{Qualitative Error Analysis}
\label{subsec:error-analysis}

To provide further insight into the experimental results, we examined the paired predictions of BanglaBERT under the context-free and flat-context settings. A representative evaluation with split seed 0 and run seed 12 was selected, in which both settings were tested on the same instances. The analysis focused on cases in which contextual information corrected or introduced prediction errors, as well as errors that remained under both settings.

We first examined errors corrected by contextual information. For example, the comment \includegraphics[height=1.7ex]{figs/bengali-text/vua.png} (fake) was incorrectly predicted as not bully without context but correctly classified as bully when its video caption was included. Similarly, the reply ``Right'' was misclassified as not bully in isolation, while the parent comment provided the conversational context needed for the correct bully prediction. Context also corrected a false positive for \includegraphics[height=1.7ex]{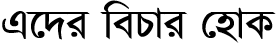} (They should be brought to justice), which was incorrectly predicted as bully without context but correctly classified as not bully when the associated caption was considered. These examples show that contextual information can resolve ambiguity in context-dependent expressions.

We next examined cases in which contextual information introduced an error. For example, the comment \includegraphics[height=1.7ex]{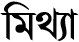} (false) was correctly classified as not bully in the context-free setting but misclassified as bully when the associated caption was included. Conversely, \includegraphics[height=1.7ex]{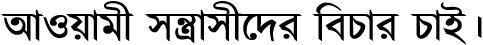} (Awami terrorists should be brought to justice) was correctly classified as bully without context but misclassified as not bully in the flat-context setting. These cases show that contextual information can sometimes introduce misleading cues or weaken discriminative information in the target text.

We also identified cases that remained false negatives under both settings, where instances labeled as bully were classified as not bully with and without context. For example, ``VONDO'' (hypocrite) and \includegraphics[height=1.8ex]{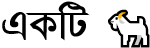} (a goat) were labeled as bully but predicted as not bully under both settings. Similarly, ``Jann0r'' (animal) remained misclassified even after contextual information was included. These cases suggest that implicit insults, Romanized Bengali expressions, and symbolic forms can remain challenging even when conversational context is available.

Conversely, some instances labeled as not bully were classified as bully under both settings, resulting in false-positive errors. For example, \includegraphics[height=1.7ex]{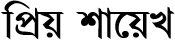} (Dear Sheikh), \includegraphics[height=1.7ex]{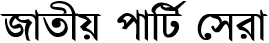} (Jatiya Party is the best), and \includegraphics[height=1.7ex]{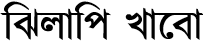} (I will eat jalebi) remained incorrectly classified as bully under both settings. These cases indicate that benign or supportive expressions within politically charged discussions can still be difficult to distinguish from bullying content.

\section{Potential Application Tasks}
\label{sec:application}

The \textit{BenCoDiC} corpus supports several potential research applications in cyberbullying analysis:

\begin{itemize}
\item \textbf{Cyberbullying Detection (\textit{Task A}):} Using the \textit{TextualContent} and \textit{Context} dimensions to develop models that classify textual instances as \textit{bully} or \textit{not bully}. The influence of background conversational context on this task is evaluated in Section~\ref{sec:experiment}.

\item \textbf{Perpetrator Identification (\textit{Task B}):} Leveraging persistent anonymized user identifiers and labeled interaction histories to identify users who repeatedly generate bullying content and to examine their activity patterns.

\item \textbf{Behavioral and Network Analysis (\textit{Task C}):} Integrating the \textit{User}, \textit{Video}, \textit{Context}, \textit{Time}, \textit{ImpactScore}, \textit{PoliticalParty}, \textit{Language}, and \textit{TargetIdentity} dimensions to analyze perpetrator behavior, targeting patterns, engagement characteristics, and community-level interactions.

\item \textbf{Link Prediction among Perpetrators (\textit{Task D}):} Constructing perpetrator networks in which nodes represent anonymized users and edges capture behavioral or contextual similarities derived from multiple dimensions. Such networks can support the prediction of potential associations among perpetrators.

\end{itemize}

Beyond these tasks, the multidimensional metadata in BenCoDiC also provides opportunities for metadata-aware multitask and few-shot learning. Auxiliary information such as language category, target identity, political-party information, temporal attributes, and engagement characteristics can be explored jointly with cyberbullying classification in future studies.

\section{Conclusion}
\label{sec:conclusion}

This work presents the Bengali Contextual and Diverse Cyberbullying (\textit{BenCoDiC}) corpus, a publicly available resource comprising 50,250 textual instances from Bengali political news discourse. The corpus preserves caption--comment--reply relationships through a nine-dimensional star schema integrating users, videos, textual content, context, time, engagement, political parties, target identities, and language categories. A two-tier human annotation process with expert verification achieved Fleiss' $\kappa=0.86$ and Krippendorff's $\alpha=0.86$. Experimental results further show that incorporating conversational context yields modest but consistent improvements over isolated-text classification.

\textit{BenCoDiC} supports context-aware cyberbullying detection and multidimensional research on perpetrator behavior and interaction networks. Future work will expand the corpus beyond political news, improve the representation of underrepresented language categories, align the resource with FAIR and Linked Data principles, and investigate ontology- and graph-based methods for perpetrator identification, behavioral and network analysis, and link prediction.

\section*{Limitations}

This study focuses on political news collected from selected Bengali YouTube channels during a specific period and may therefore exhibit topical, temporal, and selection bias. The corpus should primarily be interpreted as a resource for Bengali political-news discourse rather than as a representative sample of general Bengali social-media communication. Comments constitute approximately 88.3\% of the corpus, while political-category, target-identity, and language distributions remain domain-specific. Bengali-script text dominates the corpus, whereas Romanized Bengali, Bengali--English code-mixed text, English, and emoji-only content are underrepresented. The class distribution is also moderately imbalanced, with approximately 60\% \textit{Bully} and 40\% \textit{Not Bully} samples.

The auxiliary political-party, target-identity, and language attributes were treated as metadata and were not subjected to the two-tier inter-annotator reliability assessment used for the binary cyberbullying labels. The preliminary AI-chatbot evaluation was limited to 500 samples and free-tier systems. In addition, because the Level-2 experts had access to the Level-1 primary labels during verification, the expert review process may be susceptible to anchoring bias. These limitations should be considered when interpreting and reusing the corpus.






\bibliographystyle{IEEEtran} 
\bibliography{reference}

\end{document}